%% file: gaze.tex
\documentclass[review]{elsarticle}
\usepackage[normalem]{ulem}
\usepackage{subcaption}
\usepackage{lineno}
\usepackage{bbm}
\usepackage{amsmath}

\modulolinenumbers[5]
\usepackage[table,dvipsnames]{xcolor}
\usepackage{color}
\usepackage{diagbox}
\usepackage{slashbox}
\usepackage{multirow}
\usepackage{amsfonts}
\usepackage{bigstrut}
\usepackage{subcaption}
\usepackage{soul}
\usepackage{tabularx}
\usepackage{xcolor}
\usepackage{booktabs}
\usepackage{makecell}
\usepackage{adjustbox}
\usepackage[super]{nth}

\colorlet{tdcolor}{yellow!35}
\sethlcolor{tdcolor}
\definecolor{bluencs}{rgb}{0.0, 0.53, 0.74}
\definecolor{bleudefrance}{rgb}{0.19, 0.55, 0.91}
\journal{Pattern Recognition}
\usepackage{gensymb}
\usepackage{subcaption}
\usepackage{lineno}
\usepackage{algorithm}
\usepackage{algpseudocode}
\usepackage{tikz}
\usepackage{comment}
\usepackage{bbm}
\usepackage{multirow}
\usepackage{array}
\usepackage{tabu}
\usepackage{multicol, graphicx, rotating}
\usepackage{makecell, threeparttable, caption}
\usepackage{array}
\newcolumntype{L}[1]{>{\raggedright\let\newline\\\arraybackslash\hspace{0pt}}m{#1}}
\newcolumntype{C}[1]{>{\centering\let\newline\\\arraybackslash\hspace{0pt}}m{#1}}

\usepackage{float}
\usepackage{pifont}
\usepackage{graphicx}
\usepackage{amsmath, amsfonts, amssymb}
\usepackage{array}
\usepackage{booktabs}
\usepackage{url}
\usepackage{algpseudocode}
\usepackage{makecell}
\usepackage{tikz}
\usepackage{comment}
\usepackage{setspace}
\modulolinenumbers[5]
\usepackage{color}
\usepackage{diagbox}
\usepackage{slashbox}
\usepackage{multirow}
\usepackage{amsfonts}
\usepackage{bigstrut}
\usepackage{subcaption}
\usepackage{soul}
\usepackage{tabularx}
\usepackage{xcolor}
\usepackage{booktabs}
\usepackage{makecell}
\usepackage{adjustbox}
\usepackage{rotating}
\usepackage[super]{nth}
\usepackage{wrapfig}
\colorlet{tdcolor}{yellow!35}
\sethlcolor{tdcolor}
\definecolor{bluencs}{rgb}{0.0, 0.53, 0.74}
\definecolor{bleudefrance}{rgb}{0.19, 0.55, 0.91}
\journal{Pattern Recognition}

\begin{document}
\begin{frontmatter}

\title{Appearance Debiased Gaze Estimation via Stochastic Subject-Wise Adversarial Learning} 

% \tnoteref{mytitlenote}}

\author[1]{Suneung~Kim}
\ead{se\_kim@korea.ac.kr}
% \author[2]{Gun-Hee~Lee}
% \ead{gunhlee@korea.ac.kr}
\author[2]{Woo-Jeoung~Nam}
\ead{nwj0612@knu.ac.kr}
\author[1]{Seong-Whan~Lee\corref{cor1}}
\ead{sw.lee@korea.ac.kr}

\affiliation[1]{organization={Department of Artificial Intelligence, Korea University},
            addressline={Anam-dong, Seongbuk-gu}, 
            city={Seoul},
            postcode={02841},
            country={Republic of Korea}}
            
% \affiliation[2]{organization={Department of Computer and Radio Communications Engineering, Korea University},
%             addressline={Anam-dong, Seongbuk-gu}, 
%             city={Seoul},
%             postcode={02841},
%             country={Republic of Korea}}

\affiliation[2]{organization={Department of Computer Science and Engineering, Kyungpook National University},
            addressline={Daehak-ro, Buk-gu}, 
            city={Daegu},
            postcode={41566},
            country={Republic of Korea}}
            
\cortext[cor1]{Corresponding author}

\input{Writing/1_abstract}
\end{frontmatter}
\input{Writing/2_introduction}
\input{Writing/3_related}

\input{Writing/4_method}
\input{Writing/5_experiments}

\input{Writing/6_conclusion}

\bibliography{bibtex}
\end{document}

%% file: Writing/1_abstract.tex
\begin{abstract}
Recently, appearance-based gaze estimation has been attracting attention in computer vision, and remarkable improvements have been achieved using various deep learning techniques. Despite such progress, most methods aim to infer gaze vectors from images directly, which causes overfitting to person-specific appearance factors. In this paper, we address these challenges and propose a novel framework: Stochastic subject-wise Adversarial gaZE learning (SAZE), which trains a network to generalize the appearance of subjects. We design a Face generalization Network (Fgen-Net) using a face-to-gaze encoder and face identity classifier and a proposed adversarial loss. The proposed loss generalizes face appearance factors so that the identity classifier inferences a uniform probability distribution. In addition, the Fgen-Net is trained by a learning mechanism that optimizes the network by reselecting a subset of subjects at every training step to avoid overfitting. Our experimental results verify the robustness of the method in that it yields state-of-the-art performance, achieving 3.89\degree and 4.42\degree on the MPIIGaze and EyeDiap datasets, respectively. Furthermore, we demonstrate the positive generalization effect by conducting further experiments using face images involving different styles generated from the generative model.
\end{abstract}

\begin{keyword}
Appearance-based gaze estimation \sep generalization \sep adversarial loss \sep stochastic subject selection \sep meta-learning
\end{keyword}

%% file: Writing/2_introduction.tex
\section{Introduction}
Eye gaze is a crucial cue in the analysis of human behavior. Because the gaze contains abundant human intent information, it enables research on human cognition to provide insights regarding emotion and general communication \cite{andrist2014conversational,bixler2015automatic}. Human gaze estimation applies to many applications, such as virtual reality \cite{patney2016perceptually,pfeiffer2008towards}, human-computer interaction \cite{ lee2001automatic,lu2017appearance,martinikorena2019low, liu2022eye, cheng2017gazing}, gaming \cite{corcoran2012real}, content creation \cite{wedel2017review}, human-robotic interaction (HRI) \cite{andrist2014conversational,ahmad2006human,moon2014meet}, and mobile phone scenarios \cite{krafka2016eye}. With the exploitation of deep learning techniques, appearance-based gaze estimation, which directly maps an image to a person’s gaze, has recently received significant attention in computer vision. This technology makes it possible to learn high-level gaze features from high-dimensional images, which results in significant improvements in the accuracy of appearance-based gaze estimation methods. To improve gaze accuracy, many researchers have presented large-scale datasets to assist in gaze studies \cite{krafka2016eye,zhang2017mpiigaze,zhang2017s,zhang2020eth}.

However, as appearance-based gaze estimation is a direct regression method, consistent gaze prediction is difficult owing to the additional information contained in images, such as illumination, occlusion, decoration,  and person-specific factors. Among these, person-specific appearance factors induce overfitting because datasets are created with a limited number of participants. Therefore,  applying gaze estimation to real-life applications remains a challenging problem.

To address this problem, various gaze estimation methods have recently been proposed that use deep learning mechanisms, such as adversarial training \cite{wang2019generalizing} and few-shot learning \cite{park2019few}. The adversarial method \cite{wang2019generalizing} attempts to extract the person-invariant features from the left and right eyes of a person separately; however, it is unclear how it improves generalization, and there are no evaluation metrics that measure generalization performance. The few-shot method \cite{park2019few} introduces a geometric constraint on gaze representations and disentangles the appearance factor, gaze, and head pose to personalize these data. However, this study focuses on personal calibration samples in datasets with similar environments, and requires test-subject-specific labels for generalization. To improve robustness, various appearance-based gaze estimation methods have been proposed \cite{biswas2021appearance,yu2018deep,xiong2019mixed,cheng2018appearance,liu2019differential}. The gaze method \cite{biswas2021appearance} presents attention modules, and the multitask-learning \cite{yu2018deep} simultaneously infers eye landmarks and gaze vectors. Unfortunately, these methods focus on robustness rather than generalization and do not provide satisfactory solutions to the appearance generalization problem.

In this work, we propose a novel framework, Stochastic subject-wise Adversarial gaZE learning(SAZE), to address the problem of appearance generalization. Fig. 1 shows the overview of the SAZE framework, which focuses on how to map the face image to gaze. To learn a generalized gaze representation for appearance, the SAZE uses a Face generalization Network (Fgen-Net) consisting of a face-to-gaze encoder and a face identity classifier and utilizes the following two strategies for the network training.

First, Fgen-Net is trained using an adversarial strategy consisting of two steps to generalize the face appearance factors. In the first step, the face identity classifier is updated to distinguish face identities. In the next step, the face-to-gaze encoder is updated through general gaze loss and proposed adversarial loss. The proposed loss allows the face identity classifier to infer a uniform probability distribution for the face identity of all subjects in the training data (thus maximizing the entropy for face identities). Through this approach, it is possible that the face-to-gaze encoder predicts appearance-independent gaze features for any subject.

Second, to mitigate overfitting caused by the limited number of subjects in a dataset, we construct a new learning strategy inspired by meta-learning (learning for learning)  \cite{nichol2018first}. To obtain a generalized model for appearance, we train the Fgen-Net with meta-training and meta-adapting subsets consisting of non-overlapping subjects, which are to be reconstructed at every learning stage. Through it, we prevent the model from overfitting to specific subjects and even improve the gaze accuracy.

The experimental results demonstrate that SAZE achieves state-of-the-art performance on the MPIIGaze and EyeDiap datasets using standard evaluation metrics. Furthermore, we present a new generalization performance evaluation indicator and demonstrate the effectiveness of the proposed framework. For the purpose of evaluating generalization, we employ a generative model \cite{shen2020interpreting} with the ability to transform an unconditionally trained face synthesis model into a controllable generative adversarial networks (GAN). The model enables facial editing, encompassing aspects such as pose, age, expression, eyeglasses, and more, through the interpretation of the initial latent space and the exploration of underlying semantic subspaces. The generative model is used to generate sets of facial images exhibiting diverse styles yet similar gaze directions. Subsequently, we analyze the generalization performance of the trained gaze model by examining the gaze direction vectors inferred from these generated facial images.

\begin{figure}
\centering
\includegraphics[height=5.7cm]{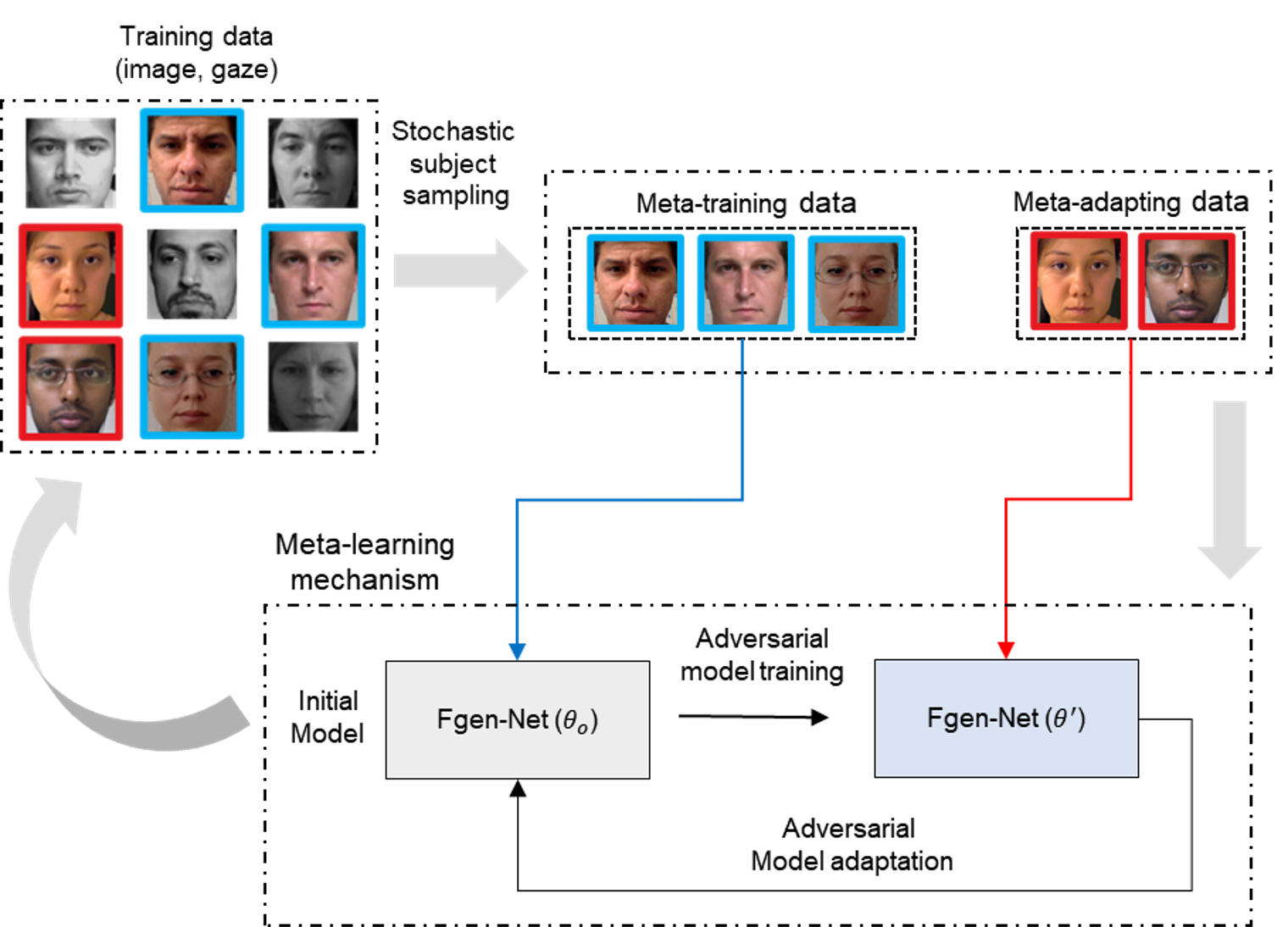}
\caption {Overview of the SAZE framework. Given training images with ground-truth gaze directions, we first select subjects in the training data and then construct meta-training and meta-adapting sets. The Face generalization Network (Fgen-Net) is trained using the proposed adversarial loss and meta-learning using the two composed subsets.}
\label{fig:example}
\end{figure}

To summarize, our main contributions are as follows:

\begin{enumerate}
    \item We propose adversarial training for gaze estimation that generalizes the appearance of subject’s faces. This allows the face identity classifier to predict uniform probabilities for all subjects, making it possible to extract appearance-independent gaze representations.
    \item To improve generalization performance, we apply a novel stochastic subject-wise training strategy. This prevents the model from being overfitted to a specific subject appearance by constructing subsets of non-overlapping subjects and training the model using meta-learning.
    \item The proposed framework achieves state-of-the-art performance of 3.89$\degree$ and 4.42$\degree$  on the MPIIGaze and Eyediap datasets, respectively. We further demonstrate the generalization effectiveness through a new evaluation indicator that analyzes the distribution of gaze for images generated from a generative model.
\end{enumerate}

%% file: Writing/3_related.tex
\section{Related Work}

\subsection{Appearance-Based Gaze Estimation}

The appearance-based gaze estimation method maps captured images using a general commodity camera directly through a network into a gaze direction vector without requiring handcrafted features.

Zhang et al. \cite{zhang2017s} presented the first appearance-based method for eye images using a CNN-based neural network designed using LeNet. Since then, the performance of gaze estimation has been significantly improved through various methods, such as a multitasking model for gaze estimation using the landmark constraint of the eye \cite{yu2018deep}, a method applying a coarse-to-fine strategy \cite{cheng2020coarse}, and an attention mechanism for gaze estimation \cite{biswas2021appearance}. However, these methods only consider the accuracy performance using public datasets \cite{krafka2016eye,zhang2017mpiigaze,zhang2017s,zhang2020eth} and ignore the appearance generalization problem. Recently, Bayesian approaches \cite{wang2019generalizing} and few-shot learning methods \cite{yu2019improving,park2019few} have been used to propose new learning mechanisms and novel architectures to address this generalization problem; however, these methods focus on personal calibration samples from datasets with similar environments (i.e., if there are no labels for test subjects, effective generalization cannot be expected). Here, we propose a method that essentially allows the gaze estimation network to learn to generalize appearance using only the images of subjects contained in the training dataset.

\subsection{Domain Generalization}

Domain generalization is a challenging problem because it requires generalizing unseen domains in a model without knowledge of their distribution during training. The key to address this problem is to extract task-specific features that are domain-invariant \cite{li2018deep,li2018domain,ghifary2015domain,muandet2013domain, shi2023source, cheng2023adversarial}. 

Muandet et al. \cite{ghifary2015domain} proposed a domain-invariant component analysis algorithm that minimizes the dissimilarities across domains. Ghifary et al. \cite{muandet2013domain} proposed a multitask autoencoder method to learn domain-invariant features. Li et al. \cite{li2018domain} introduced a generalization method that minimizes the discrepancies in a joint distribution by considering the conditional distribution of a label space extracted from input data. Recently, adversarial techniques have become a hot topic, and many studies on domain adaptation and generalization using adversarial approaches have been proposed. Li et al. \cite{li2018deep} aligned the multi-domain distribution using an extended adversarial autoencoder, and Tzeng et al. \cite{tzeng2017adversarial} presented an adversarial training method for domain adaptation. Shi et al. \cite{shi2023source} achieved comparable generalization performance and higher data efficiency without transferring of source and target models using distributionally adversarial training. Cheng et al. \cite{cheng2023adversarial} proposed a method to regularize the distributions of different classes using adversarial training. Furthermore, in the utterance domain, voice conversion (VC) using an adversarial strategy was presented \cite{zhang2019non}. Zhang et al. \cite{zhang2019non} designed an adversarial training strategy to generalize speaker-related domain information from a linguistic representation and achieved efficient voice conversion. Inspired by this method, we propose a novel framework to generalize appearance-related information from gaze representations.

\subsection{Meta-Learning}

Meta-learning described as ``learning to learn", is a long-standing topic in deep learning that quickly and effectively optimizes model parameters based on prior experience. Gradient-based methods \cite{nichol2018first,finn2017model} are representative meta-learning techniques that have been applied in many deep learning tasks. 

MAML \cite{finn2017model} applies an increasing gradient update step in the base model using a first-order approximation. This reduces the expensive computational costs required for the second-order gradient update step, which improves the speed of the learning process and generalization performance. Hence, several studies have addressed domain generalization by applying the training paradigm used by MAML \cite{li2018learning,balaji2018metareg,ye2021novel, xu2021unsupervised}. Reptile \cite{nichol2018first} is similar to MAML but only uses first-order gradients. We used this concept in our training strategy to improve the performance of face appearance generalization.

%% file: Writing/4_method.tex
\section{Proposed Approach}
\begin{figure*}[h]
\centering
\includegraphics[height=10cm, width=12cm]{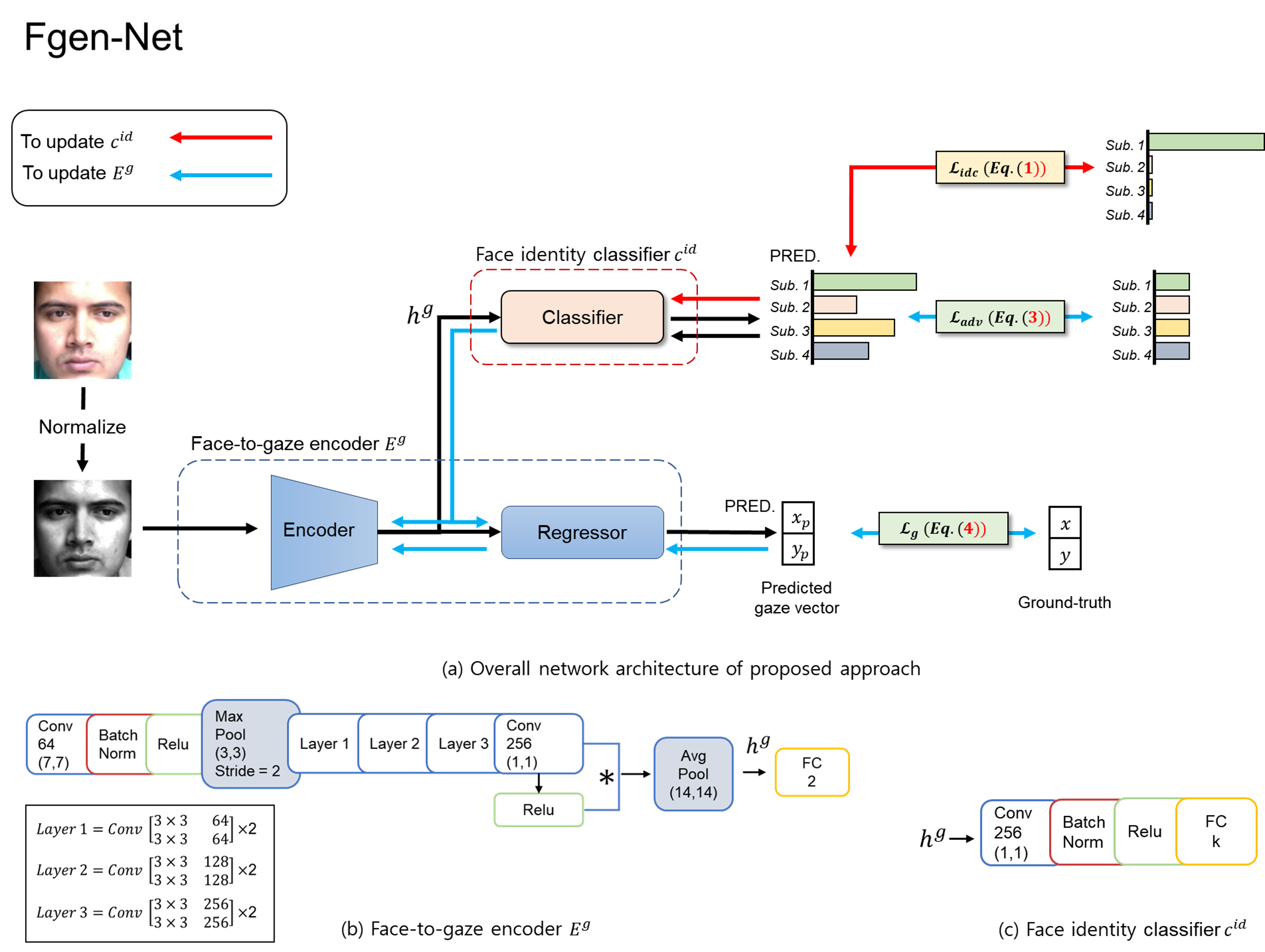}
% \end{center}
  \caption{Illustration of Face generalization Network (Fgen-Net) learning based on adversarial loss. (a) Fgen-Net is optimized through two forward paths. In the first path, the face identity classifier $ c^{id}$ is updated through the identity loss $ L_{idc}$ (Eq. (1)), which distinguishes the face appearance. Next, in the second path, the face-to-gaze encoder $ E^{g}$ is updated through the adversarial loss $ L_{adv}$ (Eq. (3)) and gaze loss $ L_{g}$ (Eq. (4)) and the adversarial loss induces the face classifier to predict a uniform probability for face appearances, which achieves a better generalization performance. (b) Detailed architecture for face-to-gaze encoder $ E^{g}$. \textbf{*} means element-wise multiplication. (c) Detailed architecture for face identity classifier $ c^{id}$. }
\label{fig:short}
\end{figure*}

In this section, we introduce our proposed Stochastic subject-wise Adversarial gaZE learning (SAZE) approach. Fig. 2 presents an overview of the Face generalization Network (Fgen-Net) used in our framework. The main purpose is to learn gaze representation generalized to face appearance through an adversarial loss using a face identity classifier and to improve generalization performance using a subject-wise meta-learning technique. We first provide an architecture overview of the SAZE framework and then describe the proposed technologies in detail.

\subsection{Architecture Overview}

The Face generalization Network (Fgen-Net) incorporates two components: a face-to-gaze encoder $\boldsymbol{E^{g}}$ and a face identity classifier $\boldsymbol{c^{id}}$ (Fig. 2). We use SWCNN \cite{zhang2017s} as the baseline model and extend the network architecture by concatenating a face identity classifier to achieve appearance-independent gaze estimation. Details follow.

\textbf{face-to-gaze encoder} $\boldsymbol{E^{g}}$: The face-to-face encoder converts face images into gaze direction vectors (yaw and pitch) as $(\boldsymbol{\hat{v}^{g}$, $\boldsymbol{h^{g}})$ $= E^{g}(D)}$ , where $\boldsymbol{D}$ is the normalized face image set and $\boldsymbol{\hat{v}^{g}}$ is gaze vectors and $\boldsymbol{h^{g}}$ is representation extracted from the middle layer.

\textbf{face identity classifier} $\boldsymbol{c^{id}}$: The face identity classifier receives the gaze representation $\boldsymbol{h^{g}}$ vector as input, which is obtained before predicting the gaze direction vector, and then predicts the face identity probability $\boldsymbol{\hat{p}^{id}}$ $= [\hat{p}_{1}^{id},...\hat{p}_{k}^{id}]$ for the number of subjects $k$ as $\boldsymbol{\hat{p}^{id} = c^{id}(h^{g})}$.

The proposed architecture, Fgen-Net, is designed to extract a gaze representation $\boldsymbol{h^{g}}$ generalized to an appearance by predicting a uniform probability value for all faces.

\subsection{Appearance-Invariant Feature Learning}
In this section, we describe the adversarial training process. To obtain a face-to-gaze encoder generalized for face appearance factors, Fgen-Net uses two forward propagation paths.

In the first path (to update $\boldsymbol{c^{id}}$ in Fig 2), the identity probability $\boldsymbol{\hat{p}^{id}}$ for k subjects is predicted by the face identity classifier  $\boldsymbol{c^{id}}$. The model parameters of the face identity classifier are then updated through the cross-entropy loss on the identity probability and face identity label. At this time, the model parameter of the face-to-gaze encoder $\boldsymbol{E^{g}}$ is frozen (i.e., not updated). The face identity loss is defined as

\begin{equation}\label{eu1}
    L_{idc} = \frac{1}{N}\sum_{n=1}^N CE(p^{id}, \hat{p}_n^{id}),
\end{equation}
where $\boldsymbol{p^{id}}$ is a one-hot face identity label for each subject, $\boldsymbol{\hat{p}_{n}^{id}}$ is the predicted identity probability of the $n^{th}$ subject, and N is the total number of training data.

Next, Fgen-Net proceeds with adversarial training, which is the main contribution of the proposed framework for inferring a gaze direction vector generalized to an appearance.
Therefore, in the second path (to update  $\boldsymbol{E^{g}}$ in Fig 2), the face-to-gaze encoder  $\boldsymbol{E^{g}}$ is optimized toward the opposite goal of the first path (i.e., the model parameters of the face-to-gaze encoder are optimized to have an equal distribution for the identity probabilities $\boldsymbol{\hat{p}^{id}}$ of all subjects). The proposed adversarial loss for appearance generalization is thus defined as

% In the second path (blue line in Fig 2), for appearance generalization, the face-to-gaze encoder $\boldsymbol{E^{g}}$ is optimized toward the opposite goal in Eq. (1), i.e., the model parameters are updated so that all identity probabilities $\boldsymbol{p^{id}}$ predicted from the face identity classifier $\boldsymbol{E^{id}}$ have equal distribution. Hence, the adversarial loss for appearance generalization was designed as

\begin{equation}\label{eu2}
    similarity(e, \hat{p}^{id}) = \frac{e \cdot \hat{p}^{id}}{\parallel e \parallel \parallel \hat{p}^{id} \parallel},
\end{equation}

\begin{equation}\label{eu3}
    L_{adv} = \frac{1}{N}\sum_{n=1}^N (1-similarity(e, \hat{p}_n^{id})),
\end{equation}
where $e = [1/k,...1/k]^T$ is a uniform distribution, and k is the number of subjects in the training set. We use cosine similarity to construct a loss function that induces the predicted identity probabilities $\boldsymbol{\hat{p}_{n}^{id}}$ to adopt a uniform distribution. Thus, the face-to-gaze encoder $\boldsymbol{E^{g}}$ is generalized based on the appearance-related information of the face images.

Finally, the gaze direction loss uses the L1 loss, defined as

\begin{equation}\label{eu4}
    L_{g} = \frac{1}{N}\sum_{n=1}^N |v^g - \hat{v}_n^g|,
\end{equation}
where $v^g$ is the ground truth of the gaze direction (yaw and pitch) and $\hat{v}_n^g$ represents the vectors predicted by the face-to-gaze encoder $\boldsymbol{E^{g}}$.

Hence, the face-to-gaze encoder is trained using a multi-objective loss function as follows

\begin{equation}\label{eu5}
    L_{total} = \lambda_{adv}L_{adv} + L_{g},
\end{equation}
where we empirically set $\lambda_{avd}=5$. When the parameters of the face-to-gaze encoder are updated, the face identity classifier is frozen. Through the above adversarial learning, we alleviate the limitations of appearance-based gaze estimation methods by generalizing appearance-related information without requiring additional test-specific labels.

\subsection{Stochastic Subject-Wise Optimizing}

To improve the generalization performance, we introduce a new learning strategy. This strategy is tailored to compensate for overfitting problems caused by person-specific appearance factors. In general, gaze estimation performs leave-one-out cross-validation, which leaves only one subject out of a test dataset, and uses the remaining subjects for training. In this training setting, the model is biased by the appearance factors of the limited subjects used in the training dataset, and overfitting easily occurs. To avoid these problems, we constructed a new training strategy that stochastically selects a subject in the training set at every training step and updates the model by applying the meta-learning method, Reptile \cite{nichol2018first}, which is a first-order gradient-based algorithm.

First, $k$ subjects are randomly selected from the training dataset, and a new meta-training set, $\boldsymbol{D_{k}^{train}}$ is constructed with these subjects. The meta-training set $\boldsymbol{D_{k}^{train}}$ is input to the proposed Fgen-Net, the gradient vectors are computed during the optimization process as follows:

\begin{equation}\label{eu6}
   U({D_{k}^{train}},\theta_o^{E^g}, m) = \theta_o^{E^g} + g_1 + g_2 + ... + g_m ,
\end{equation}
where $\theta_o^{E^g}$ is the initial weight of the face-to-gaze encoder $\boldsymbol{E^{g}}$, $m$ is the number of meta-training iterations, and $g_m$ represents the gradient vectors computed during the face-to-gaze encoder optimizing. 

Then, we update the initial weight $\theta_o^{E^g}$ using the following equations.

\begin{equation}\label{eu7}
    \theta_n^{E^g} = \theta_o^{E^g} + \epsilon(\phi - \theta_o^{E^g}),
\end{equation}
\begin{equation}\label{eu8}
    \phi =  U({D_{k}^{train}},\theta_o^{E^g}, m),
\end{equation}
where $\epsilon$ is the step size used by the stochastic gradient descent (SGD) operation and $\phi$ is the weight of the face-to-gaze encoder optimized by the meta-training process. Next, we construct a meta-adapting set $\boldsymbol{D_{p}^{adapt}}$ with p non-overlapping subjects in $\boldsymbol{D_{k}^{train}}$ and update the weight $\theta_n^{E^g}$ from the meta-adapting iteration as

\begin{equation}\label{eu9}
    U^*({D_{p}^{adapt}},\theta_n^{E^g}, j) = \theta_n^{E^g} + g_1 + g_2 + ... + g_j .
\end{equation}

After the meta-adapting iteration, we create a new meta-adapting set  $\boldsymbol{D_{p_i}^{adapt}}$ by selecting new subjects, repeat the meta-adapting process $T$ times, and calculate the updated weight $U_i^*({D_{p_i}^{adapt}},\theta_n^{E^g}, j)$.

Finally, the face-to-gaze encoder $\boldsymbol{E^{g}}$ is updated based on
\begin{equation}\label{eu10}
    \theta_{n+1}^{E^g} = \theta_{n}^{E^g} + \epsilon\frac{1}{T}\sum_{i=1}^T(\phi^*_i-\theta_{n}^{E^g}), \qquad 
\end{equation}
\begin{equation}\label{eu11}
    \phi^*_i =  U_i^*({D_{p_i}^{adapt}},\theta_n^{E^g}, j),
\end{equation}
where $\epsilon$ is same as in Eq. (7), and $j$ is the number of meta-adapting iterations, and $\phi^*_i$ is the weight of the face-to-gaze encoder optimized by the meta-adapting process using $\boldsymbol{D_{p_i}^{adapt}}$.

We applied this process to alleviate the overfitting problem, which allows the model to optimize new parameters and prevents the model from being updated in a dominant direction.

\subsection{Training Summary}
\begin{algorithm}
\caption{The SAZE training procedure}\label{alg:cap}
\begin{algorithmic}[1]
\State \textbf{Initialization:}
\State $\theta^{E^g}, \theta^{c^{id}}, iter \leftarrow 1.$
\State \textbf{Iteration:}
\For{iteration $=$ 1,2,...}
\State get meta-training set ${D_{k}^{train}}$
\State $U_o \leftarrow \theta_o^{E^g}$ 
\State computing $U_m$ during meta-training using $L_{adv}, L_{g}$ 
\State and $L_{idc}$
\For{meta-training iteration $=$ 1,2,...m}
\State $\_, h^{g} \,\,\,\ \leftarrow \quad E^g({D_k^{train}})$
\State $\hat{p}^{id} \,\,\,\,\ \leftarrow \quad c^{id}(h^{g})$
\State $\theta^{c^{id}} \ \xleftarrow{+} -\nabla_{\theta^{c^{id}}}(\lambda_{idc}L_{idc})$
\State Freezing $\theta^{c^{id}}$
\State $\hat{v}^g, \_ \,\,\,\,\,\,\ \leftarrow \quad E^g({D_k^{train}})$
\State $\theta^{E^{g}} \,\,\ \xleftarrow{+} -\nabla_{\theta^{E^{g}}}(\lambda_{avd}L_{avd} + L_g)$
\State meta-training iter $\xleftarrow{+}$ 1
\EndFor
\State update initial weight $\theta_o^{E^g}$
\State $\theta_n^{E^g} \ \leftarrow \theta_o^{E^g} + \epsilon(\phi - \theta_o^{E^g})$
\State where $\phi = U({D_{k}^{train}},\theta_o^{E^g}, m)$
\For{i $=$ 1,2,...T}
\State get meta-adapting set ${D_{p_i}^{adapt}}$
\For{meta-adapting iteration $=$ 1,2,...j}
\State perform the same process using ${D_{p_i}^{adapt}}$
\State meta-adapting iter $\xleftarrow{+}$ 1
\EndFor
\State Compute $\phi^*_i =  U_i^*({D_{p_i}^{adapt}}$,$\theta_n^{E^g}, j)$
\State i $\xleftarrow{+}$ 1
\EndFor
\State $\theta_{n+1}^{E^g} = \theta_{n}^{E^g} + \epsilon\frac{1}{T}\sum_{i=1}^T(\phi^*_i-\theta_{n}^{E^g})$
\State iter \ $\xleftarrow{+}$ 1
\EndFor
\end{algorithmic}
\end{algorithm}

To train the SAZE framework that generalizes appearance, we use three losses: the identity loss $\boldsymbol{L_{idc}}$ for distinguishing face identities, the adversarial loss $\boldsymbol{L_{adv}}$ for adversarial training, and the gaze direction loss $\boldsymbol{L_{g}}$.

Algorithm 1 outlines the entire training process, where $\boldsymbol{\theta^{E^g}}$ and $\boldsymbol{\theta^{c^{id}}}$ are the model parameters of the face-to-gaze encoder $\boldsymbol{E^{g}}$ and the face identity classifier $\boldsymbol{c^{id}}$, respectively. First, Fgen-Net is trained with the meta-training set, which consists of $k$ randomly selected subjects by applying the adversarial loss discussed in Section 3.2. Then, to apply meta-learning strategies, $U$ (Eq. 6) is computed from the results of the meta-training iteration to obtain the new weight $\boldsymbol{\theta_{n}^{E^g}}$. When the meta-training iteration is completed, we construct a meta-adapting subset $\boldsymbol{D_{p}^{adapt}}$ of non-overlapping subjects in $\boldsymbol{D_{k}^{train}}$ and update the face-to-gaze weights by repeating the previous training steps. The meta-adapting process is repeated by constructing a new meta-adapting subset, and the weight of the face-to-gaze model $\boldsymbol\theta_{n}^{E^g}$ is updated using the computed weights $\boldsymbol\phi^*_i$ (Eq. 9) obtained from the meta-adaptation process.

%% file: Writing/5_experiments.tex
\section{Implementation Details}

\subsection{Datasets}

Since the proposed framework is designed for mapping face images to gaze, we use the following two public datasets for validation. \par\par

\noindent$\textbf{MPIIGaze}$ \cite{zhang2017s} is the most popular benchmark dataset used for appearance-based methods. This dataset contains 3,000 face images of 15 subjects, which provides high within-person variations in appearance, including illumination, face shape, and hairstyle. Faces are provided in 224 × 224 pixel frontal images depending on head pose. The image pre-processing used in \cite{zhang2017s} is adopted in the experiments conducted in this study.\par

\noindent$\textbf{EyeDiap}$ \cite{funes2014eyediap} consists of 3-minute video clips involving 16 subjects. This dataset provides two target types: screen target (CS) on a fixed monitor and floating physical target (FT). For experiments, we used the screen target and applied data pre-processing to obtain images such as those appearing in the MPIIGaze dataset. As CS video data for two subjects was not present, the experiment was conducted with 14 subjects. The dataset officially available comprises approximately 6,000 to 7,000 upper body images across a total of 14 subjects, with the overall image count in the dataset totaling around 90,000. We employed the Python+Dlib face detector for extracting the facial region, and utilized the pre-processing techniques outlined in \cite{zhang2017s} for conducting our experiments.

\subsection{Training Details}

\noindent$\textbf{Architecture}$. The model used for the experiment was selected as a universal SWCNN \cite{zhang2017s} model of appearance-based gaze task as baseline, and our proposed method was applied to this model. We used a batch size of 32 and applied the Adam optimizer to both the face-to-gaze encoder and the face identity classifier. For both datasets, we set the learning rate to $10^{-3}$ and the weight decay to $10^{-3}$, which remained constant during training.\par

\noindent$\textbf{Meta process}$. During meta-training experiments, different number of subjects $k$  were selected, and the step size $\epsilon$  (Eq. 10) was set to 1 and was decreased stepwise by $1-\frac{1}{epochs}$ (Eq. 7). We set the meta training iterations $m$ to 20 steps (Eq. 6).

For the meta-adapting set, the number of subjects p was fixed at 2, and resampling $T$ was performed 5 times per epoch (Eq. 10). The meta-adapting iteration $j$ was performed 10 times for each sample (Eq. 9). The main epoch was repeated 30 times, utilizing images resized to 224x224 for model training. The training environment conditions were consistent for both the MPIIGaze and EyeDiap datasets.

\subsection{Settings to Evaluate Generalization Performance}

\begin{figure}[h]
\centering
\includegraphics[height=6.5cm]{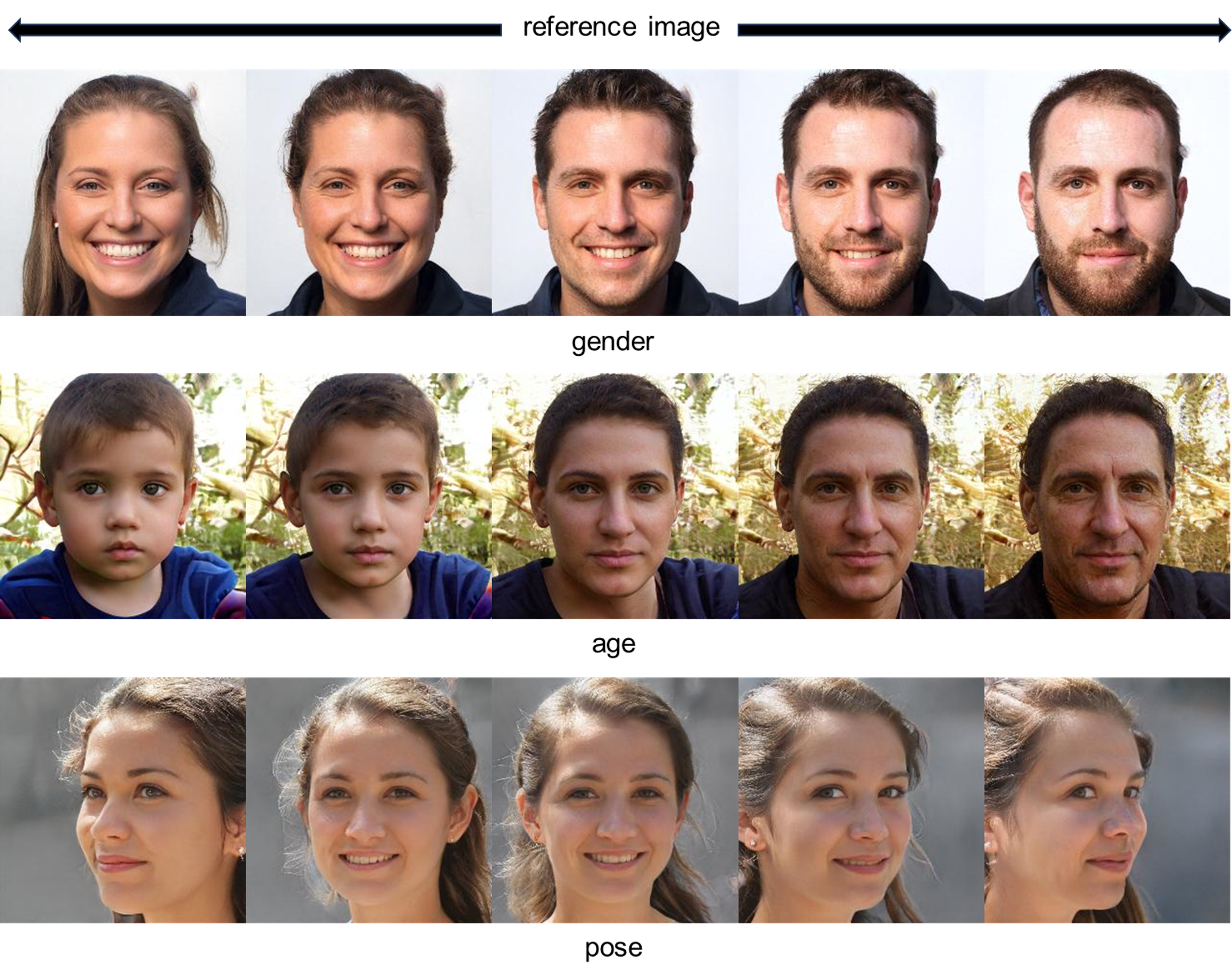}
\caption{Image examples generated by a generative model \cite{shen2020interpreting}. The first column illustrates the outcomes of gender-specific latent space adjustment, the second column pertains to age, and the third column pertains to pose.}
\label{fig:EX}
\end{figure}

Fig. 3 displays images created using the generative model \cite{shen2020interpreting}. The generative model explores the disentanglement between various semantics and manages to decouple some entangled semantics with subspace projection, leading to more precise control of facial attributes. In addition to manipulating gender, age, expression, and the presence or absence of glasses, this framework enables the adjustment of facial poses as well as the correction of artifacts accidentally produced by GAN models. Drawing inspiration from these concepts, we manipulated the latent distances of facial attributes to generate datasets exhibiting similar gazes but distinct faces, employing them to assess the generalization performance of our model. For our experiments, we curated a dataset by generating 70 images for each of the seven gaze types through latent space adjustments of age and gender. We subsequently conducted a comparative analysis of the generalization performance between the baseline model and the proposed model. Because the yaw and pitch ground truths of the constructed datasets are unknown, we evaluated the performance by analyzing the variance obtained from the gaze values.

\section{Experiment result}
\subsection{Comparison with Existing Methods}

To evaluate our proposed method, we performed leave-one-out cross-validation on the MPIIGaze and EyeDiap datasets, which selects the data of one subject as the test dataset and uses the data of the remaining subjects as the training data. We present the experimental results of the proposed approach on both datasets and compare them with existing gaze estimation approaches in Table 1. The methods used in the comparison are summarized as follows.

\begin{itemize}
    \item $\textbf{iTracker}$ \cite{krafka2016eye} performs gaze estimation by combining left/right eye images, face image information, and face grid information.
    
    \item $\textbf{SWCNN}$ \cite{zhang2017s} is a deep neural network-based gaze estimation method that learns face images by applying a spatial weighting strategy. We consider this method as our baseline model.
    
    \item $\textbf{RT-Gene}$ \cite{fischer2018rt} proposes the use of a gaze dataset in natural settings with larger camera-subject distances and less-constrained subject motion. It uses a two-stream neural network to learn two eye images.

    \item $\textbf{Dilated-Net}$ \cite{chen2018appearance} uses dilated convolutions to extract high-level eye features, which efficiently changes the receptive field size of the convolutional filters.
    
    \item $\textbf{Faze}$ \cite{park2019few} introduces a geometric constraint on gaze representations using an encoder-decoder architecture as a gaze estimation network. Thus, a highly adaptable network can be trained using only a few samples.
    
    \item $\textbf{MeNet}$ \cite{xiong2019mixed} was designed as a mixed-effect model to consider person-specific information.
    
    \item $\textbf{CA-Net}$ \cite{cheng2020coarse} uses a coarse-to-fine gaze estimation method. It first extracts facial features from a CNN network to predict the gaze direction and then uses eye features to refine the gaze direction.
    
    \item $\textbf{Bayesian Approach}$ \cite{wang2019generalizing} proposes adversarial training under semi-supervised learning to improve the gaze model’s performance on target subjects. 
    
    \item $\textbf{FAR$^*$ Net}$ \cite{cheng2020gaze} uses two eye images and face images as inputs to predict the gaze direction vectors and handles extreme head pose and illumination conditions using an evaluation-guide asymmetric regression mechanism.

    \item $\textbf{AGE-Net}$ \cite{biswas2021appearance} uses two eyes and a face as input images, and adds an Attention-branch to the feature-extraction branch of the two eyes to perform the intended feature manipulation. It enables the model to concentrate on the most pertinent information within a given context, much like how attention mechanisms aid in translation.
    
\end{itemize}

\begin{table}[h]
\begin{center}
\caption{Results of comparison with existing gaze estimation models.}
{\footnotesize
\begin{tabular}{|c|c|c|}
\hline
\multicolumn{1}{|c|}{Model} & \multicolumn{1}{|c|}{MPIIGaze} & \multicolumn{1}{|c|}{EyeDiap} \\
\hline\hline
iTracker &\multirow{1}{*}{6.2\degree} & \multirow{1}{*}{9.9\degree} \\
% \cite{krafka2016eye} & & \\

SWCNN & \multirow{1}{*}{4.8\degree} & \multirow{1}{*}{6.0\degree}\\
% \cite{zhang2017s} & & \\

RT-Gene (1 model) & \multirow{1}{*}{4.8\degree} & \multirow{1}{*}{6.4\degree}\\
% \cite{fischer2018rt} & & \\

RT-Gene (4 model) & \multirow{1}{*}{4.3\degree} & \multirow{1}{*}{5.9\degree}\\
% \cite{fischer2018rt} &&\\

Dilated-Net & \multirow{1}{*}{4.8\degree} & \multirow{1}{*}{5.9\degree}\\
% \cite{chen2018appearance} && \\

Faze  & \multirow{1}{*}{5.2\degree} & \multirow{1}{*}{-}\\
% \cite{park2019few}&&\\

MeNet & \multirow{1}{*}{4.9\degree} & \multirow{1}{*}{-}\\
% \cite{xiong2019mixed} && \\

CA-Net & \multirow{1}{*}{4.1\degree} & \multirow{1}{*}{5.3\degree}\\
% \cite{cheng2020coarse} &&\\

Bayesian Approach & \multirow{1}{*}{4.3\degree} & \multirow{1}{*}{9.9\degree}\\
% \cite{wang2019generalizing}&&\\

FAR$^*$ Net & \multirow{1}{*}{4.3\degree} & \multirow{1}{*}{5.76\degree}\\
% \cite{cheng2020gaze}&&\\

AGE-Net & \multirow{1}{*}{4.09\degree} & \multirow{1}{*}{-}\\
% \cite{biswas2021appearance}&&\\
\hline\hline
SAZE (ours)  & \textbf{3.89}\degree & \textbf{4.42}\degree\\
\hline
\end{tabular}}
\end{center}

\end{table}

SAZE achieved mean angle errors of 3.89\degree and 4.42\degree on the MPIIGaze and EyeDiap datasets, respectively. Compared to the state-of-the-art methods, our proposed method improved the performance by 4.6\% on the MPIIGaze dataset compared to AGE-Net \cite{biswas2021appearance} and improved the performance by 16.6\% on EyeDiap dataset compared to CA-Net \cite{cheng2020coarse}. In addition, SAZE only uses face images, unlike AGE-Net, which uses face and two eye images. Hence, SAZE improved the performance using less image information, thus reducing computational costs. Furthermore, compared to the SWCNN \cite{zhang2017s} baseline model, significant performance gains of 20\% and 26\% were achieved by SAZE on the MPIIGaze and EyeDiap datasets, respectively.

\subsection{Participant-Wise Analysis}

\begin{figure}[h]
\centering
\includegraphics[height=6cm, width=10cm]{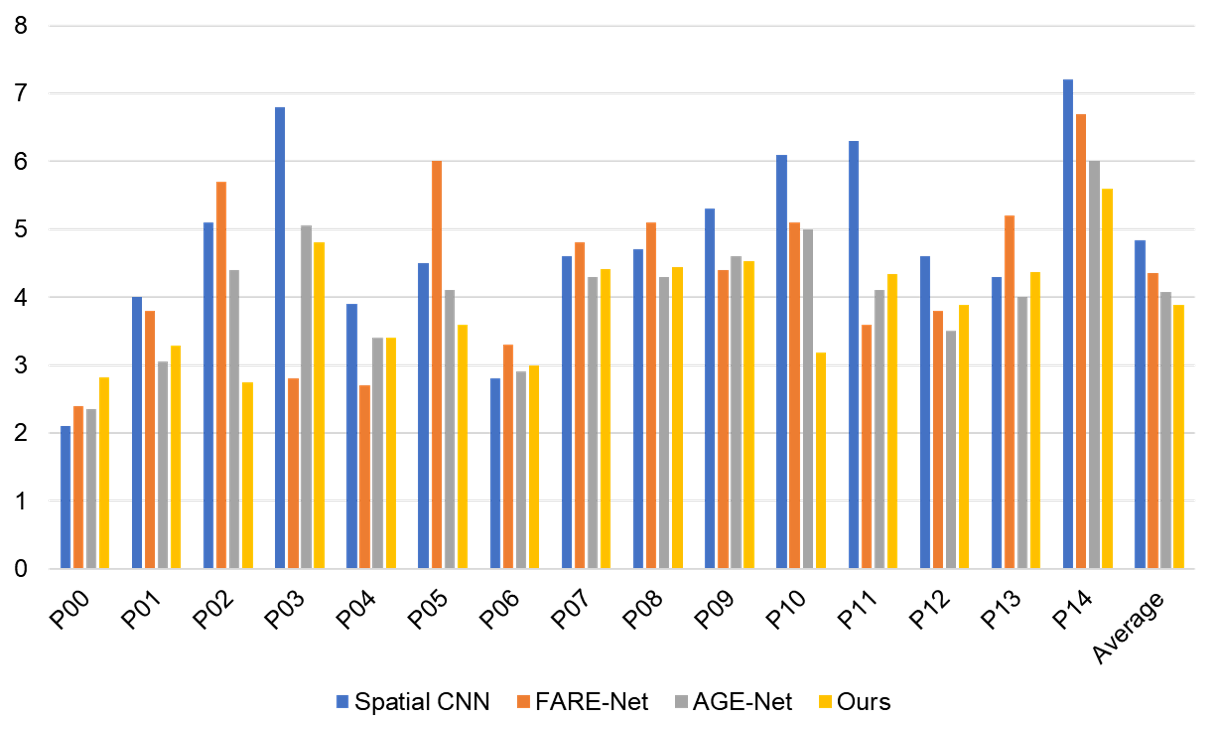}
\caption{Participant-wise gaze accuracy.}
\label{fig:example}
\end{figure}

We analyzed the performance for each of the 15 subjects in the MPIIGaze dataset by comparing the various gaze estimation methods. As shown in Fig. 4, 
our proposed method SAZE, performed better than the baseline for all subjects except p01. Furthermore, SAZE performed better for nine subjects than FAR* Net \cite{cheng2020gaze} and for seven subjects compared to AGE-Net. 
We further compared the standard deviations of the mean angle errors. When compared with AGE-Net \cite{biswas2021appearance}, the performance of SAZE was improved in seven subjects, which did not indicate an improvement over AGE-Net based on the gaze accuracy estimation; however the standard deviation was 0.79, which was 0.14 lower than that of AGE-Net . This result is interpreted as indicating a better generalization performance is achieved by SAZE.

\begin{figure}[h]

\begin{description}
\footnotesize
\item \textbf{\qquad\qquad\qquad\ MPIIGaze\qquad\qquad\qquad\qquad\qquad\qquad\qquad EyeDiap \quad } \\
\end{description}
\vspace{-0.6cm}
\centering
 \includegraphics[height = 4.7cm, width=12cm]{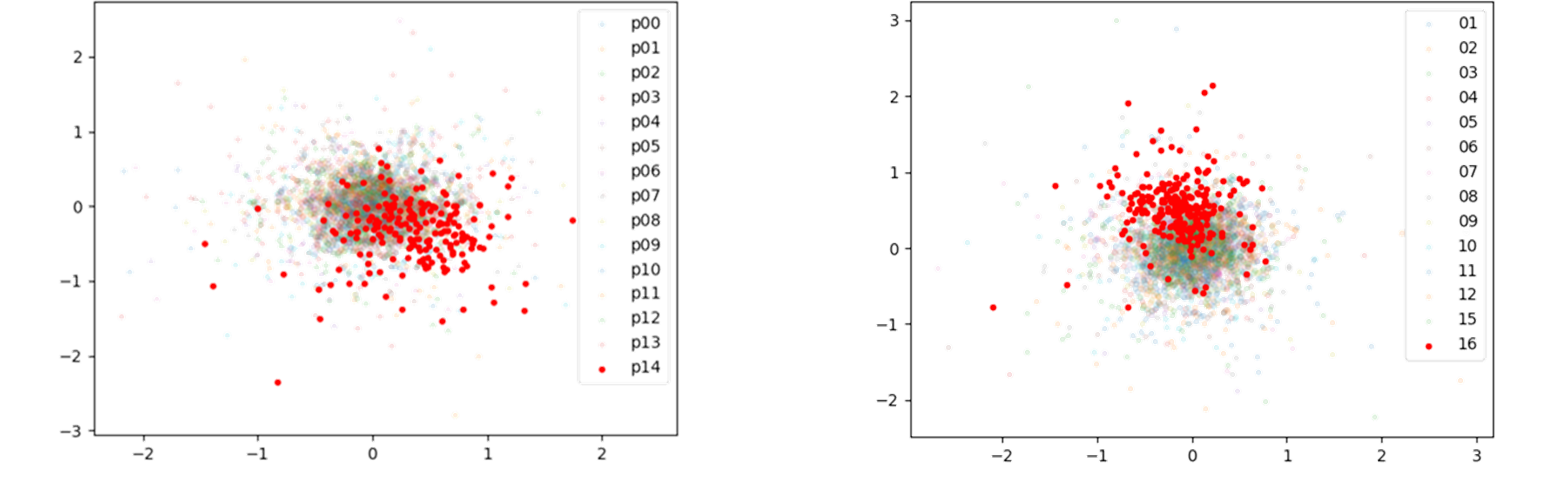}
\begin{description}
\centering
\item \textbf{(a) Distribution of t-SNE for the Spatial-Weight CNN
 (baseline)} 
\end{description}
\begin{description}
\footnotesize
\item \textbf{\qquad\qquad\qquad\ MPIIGaze\qquad\qquad\qquad\qquad\qquad\qquad\qquad EyeDiap \quad } \\
\end{description}
\vspace{-0.3cm}
\includegraphics[height = 4.7cm, width=12cm]{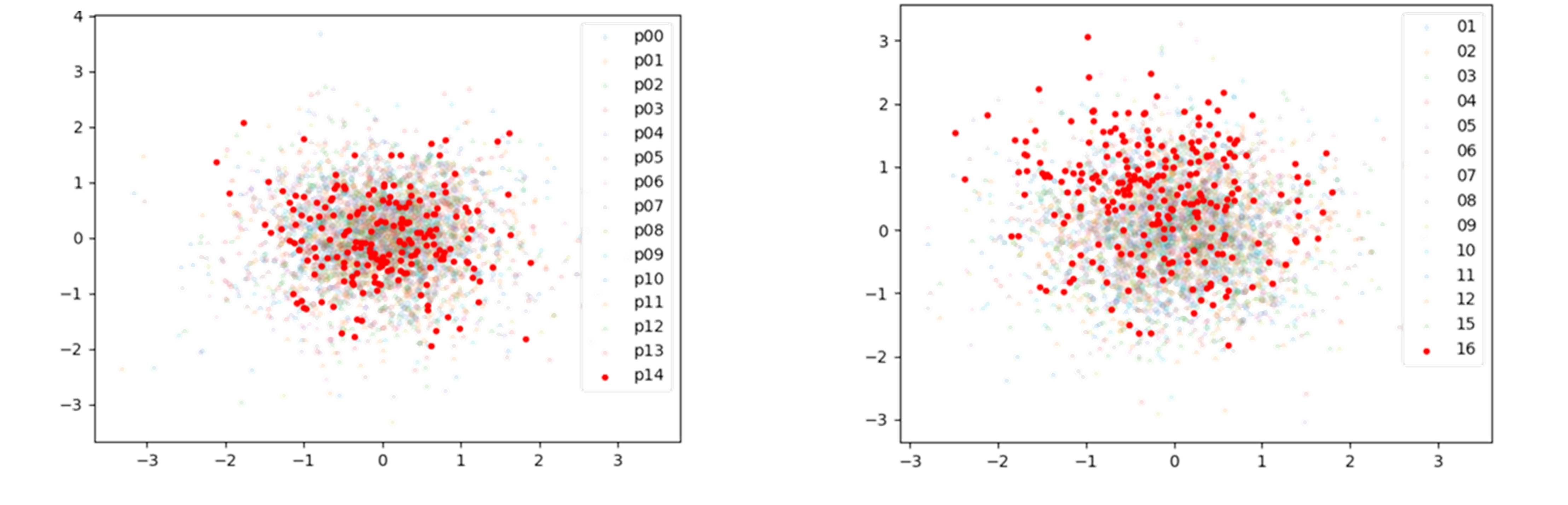}
\begin{description}
\centering
\item \textbf{(b) Distribution of t-SNE for the SAZE (ours)}
\end{description}
\vspace{-0.1cm}
  \caption{Comparison of T-SNE visualization using point cloud classification results for SWCNN and SAZE. The red points represent untrained subjects.}
\label{fig:short}
\end{figure}

\begin{figure}
% \begin{description}
% \centering
% \item \textbf{Mean Angle Error based on gaze direction values} 
% \end{description}
\centering
 \includegraphics[height = 3cm, width=12cm]{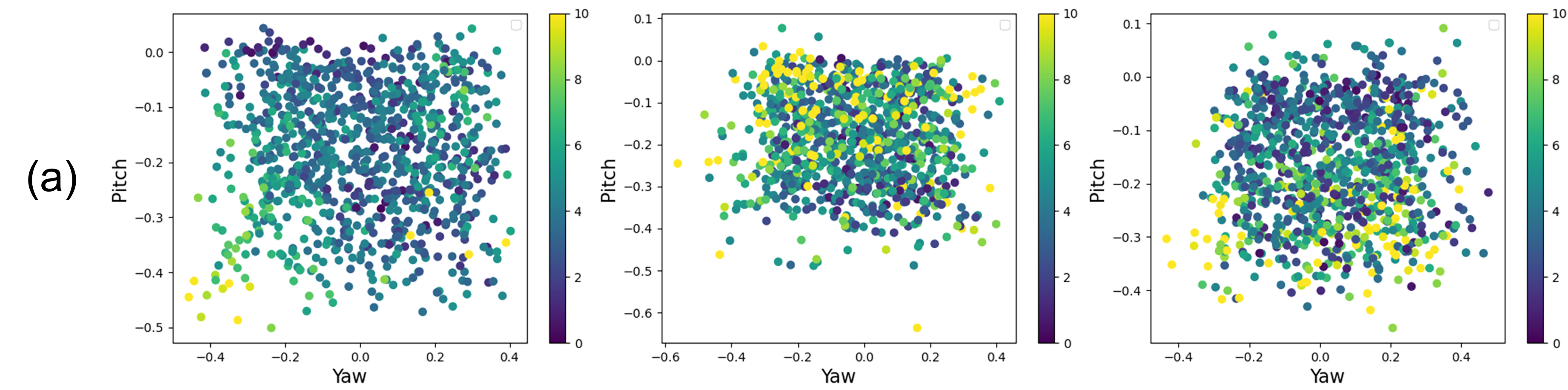}
\includegraphics[height = 3cm, width=12cm]{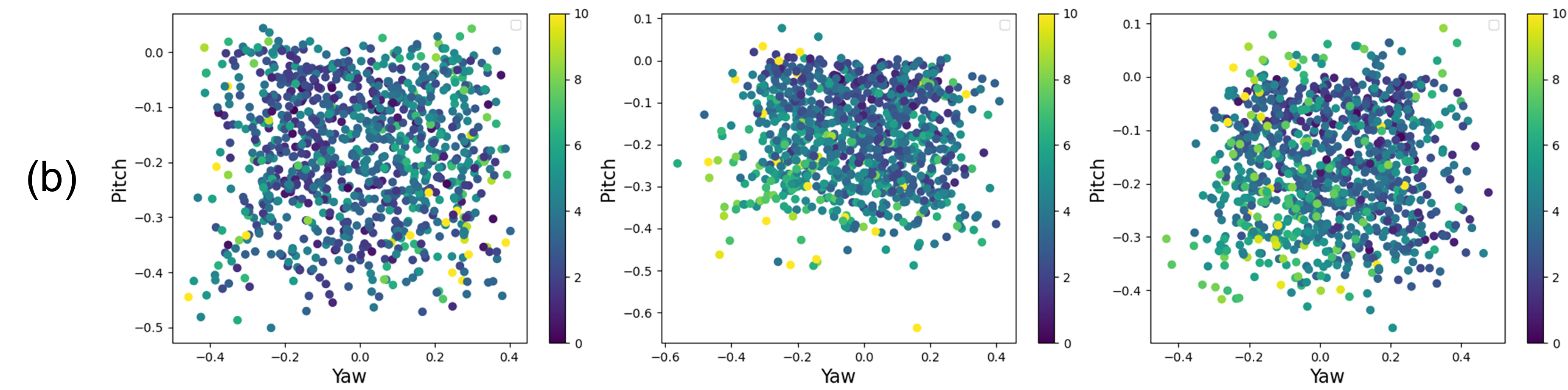}
\begin{description}
\vspace{-0.1cm}
\centering
\footnotesize
\item \textbf{MPIIGaze} \\
\end{description}
\includegraphics[height = 3cm, width=12cm]{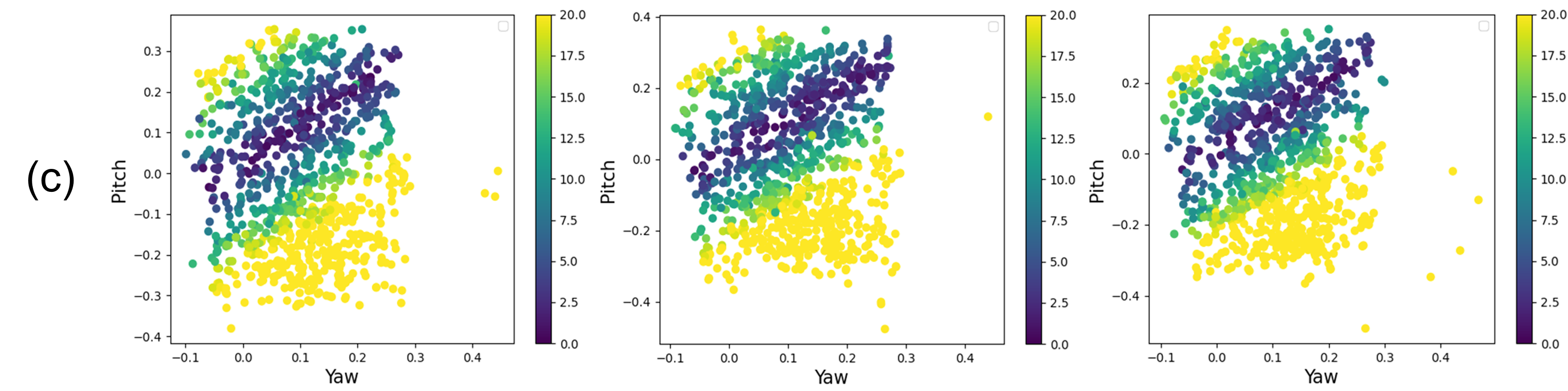}

\includegraphics[height = 3cm, width=12cm]{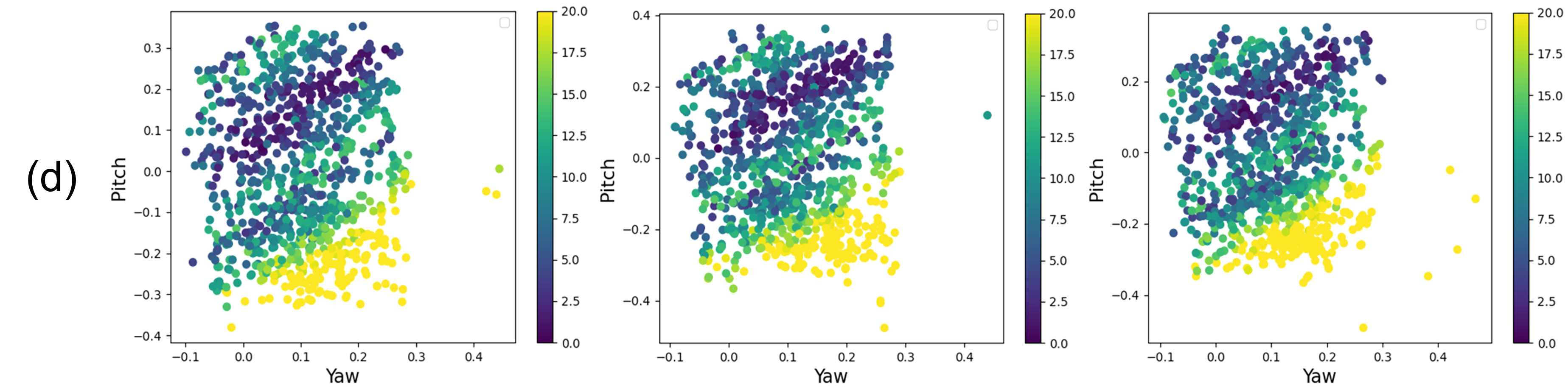}
\begin{description}
\vspace{-0.1cm}
\centering
\footnotesize
\vspace{-0.1cm}
\item \textbf{EyeDiap} \\
\end{description}
\vspace{-0.3cm}
\caption{Heat maps of the mean angle error for yaw and pitch. Figure (a) and (b) illustrate the comparison between the baseline and our method on the MPIIGaze dataset, while (c) and (d) display the results for the EyeDiap. The brighter points represent higher errors, and darker points indicate lower errors.}
\label{fig:short}
\end{figure}

\subsection{Effect of Generalization}

In order to assess the model's generalization performance, in Fig.5, we conducted an experiment where we generated  t-stochastic neighbor embedding (t-SNE) visualizations using the trained model on the subject-specific test samples. This allowed us to scrutinize the distribution patterns of both the baseline and our proposed method. For a better understanding, we compared the t-SNE distribution by extracting the gaze features of the subjects used in training and those of the untrained subjects from a model trained using both datasets. The visualization was performed using gaze features with 256 dimensions extracted after average pooling, and we implemented the algorithm using the Pytorch library. Gaze models (where the training set excluded the p14-subject and 16-subject on MPIIGaze and EyeDiap) were developed and evaluated as to how samples for untrained subjects were distributed in SWCNN as the baseline model and our proposed method SAZE. The results shown in Fig. 5(a) and 5(b) indicate the distribution achieved by SWCNN \cite{zhang2017s} and the distribution achieved by SAZE, respectively. The red points show the distribution of the data samples for untrained subjects (p14 in MPIIGaze and 16 in EyeDiap). Fig. 5(a) shows that the red points deviate from the overall distribution in both datasets for SWCNN, whereas Fig.5(b) shows that the red points are evenly distributed for SAZE. Through this experiment, we aim to confirm that for the untrained subject set, gaze values exhibit a wide distribution across diverse gaze directions, without being biased or concentrated in specific regions. Hence, these findings suggest that our model demonstrates a higher degree of convergence to the training data distribution when applied to untrained data; in other words, our method achieves better generalization performance. In addition, the variances of the red points using SWCNN are 0.1703 and 0.1244 for MPIIGaze and eyeDiap, respectively, and the variance values using SAZE are 0.6820 and 0.7602, respectively. This quantitative evaluation result shows that our method does not bias the appearance information of the subjects compared to the base model.

For more in-depth analysis, we calculated the average angular errors for yaw and pitch using untrained samples from both the MPIIGaze and EyeDiap datasets, comparing the baseline with our method. In the case of the MPIIGaze dataset, we divided the subjects into three groups: 1-5, 6-10, and 11-15, each serving as the test set. After training the model, we measured the mean angle errors on the respective test sets, each comprising 1,000 samples. For the EyeDiap dataset, we used subjects 14, 15, and 16 as the test set. Following model training, we randomly selected 3,000 samples to measure the mean angle errors. Fig. 6 presents the heat map results for the mean angle errors, which can be easily observed using the color bar on the right.
Upon reviewing the results from the MPIIGaze dataset, it is evident that the baseline (Fig. 6(a)) exhibits numerous brighter points, indicating higher error, in comparison to our method (Fig. 6(b)). Shifting focus to the EyeDiap dataset, we observe that the baseline (Fig. 6(c)) displays a significant region of elevated error, particularly for pitch values less than 0. In contrast, our method (Fig. 6(d)) effectively mitigates these errors. In conclusion, we noticed a higher number of samples exhibiting a low angle error in the proposed model compared to the SWCNN model used as the baseline. Such comparison indicates that our proposed framework is better and demonstrates that our method improves the generalization performance as well as gaze accuracy.
% using SWCNN increases as the pitch value increases (brighter points), whereas Fig. 6(b) indicates that our method does not exhibit this increase. In Fig. 6(b), SAZE produces a high density of errors below 5\degree, whereas in Fig. 6(a), SWCNN produces a high density of errors above 6\degree, based on the color bar. Such comparison indicates that our proposed framework is better and demonstrates that our method improves the generalization performance as well as gaze accuracy.

% using SWCNN increases as the pitch value increases (brighter points), whereas Fig. 6(b) indicates that our method does not exhibit this increase. In Fig. 6(b), SAZE produces a high density of errors below 5\degree, whereas in Fig. 6(a), SWCNN produces a high density of errors above 6\degree, based on the color bar. Such comparison indicates that our proposed framework is better and demonstrates that our method improves the generalization performance as well as gaze accuracy.

\subsection{Ablation Study}

\begin{table}[h]
\begin{center}
\caption{Ablation analysis. ‘adv’ denotes our proposed adversarial loss and ‘stochastic subject selection’ denotes the method presented in Section 3. As can be observed, each proposed component plays an important role and contributes to gaze estimation performance.}
{\footnotesize
\begin{tabular}{ccccc}
\cmidrule{1-1} \cmidrule{2-5}
\multirow{2}{*}{}                                                              & \multicolumn{4}{c}{Datasets}                               \\ \cmidrule(lr){2-3}\cmidrule(lr){4-5}
                                                                               & \multicolumn{2}{c}{MPIIGaze} & \multicolumn{2}{c}{EyeDiap} \\ \cmidrule(lr){2-3}\cmidrule(lr){4-5} 
                                                                               & w/o adv    & w adv           & w/o adv   & w adv           \\ \cmidrule(lr){1-1}\cmidrule(lr){2-2}\cmidrule(lr){3-3}\cmidrule(lr){4-4}\cmidrule(lr){5-5} 
\begin{tabular}[c]{@{}c@{}}w/o Stochastic\\ subject selection\end{tabular}     & 4.8\degree        & 4.03\degree            & 6.0\degree       & 5.02\degree            \\ \midrule
\begin{tabular}[c]{@{}c@{}}train subject k=5\\ adapt subject p=2\end{tabular}  & 4.47\degree       & 4.08\degree            & 4.73\degree      & \textbf{4.42\degree}   \\ \hline
\begin{tabular}[c]{@{}c@{}}train subject k=8\\ adapt subject p=2\end{tabular}  & 4.33\degree       & \textbf{3.89\degree}   & 4.91\degree      & 4.51\degree            \\ \hline
\begin{tabular}[c]{@{}c@{}}train subject k=10\\ adapt subject p=2\end{tabular} & 4.42\degree       & 3.97\degree            & 5.12\degree      & 4.73\degree            \\ \hline
\end{tabular}}
\end{center}
\end{table}

We performed ablation studies on both datasets to confirm the effects of adversarial loss and stochastic subject selection. Table 2 summarizes the overall ablation results. 

In the ablation study, the performance improved in all four cases after applying stochastic-wise optimization on the MPIIGaze and EyeDiap datasets. In addition, without subject-wise optimization, the gaze accuracy performance improved by 16\% on MPIIGaze (4.03\degree) and 17\% on EyeDiap (5.02\degree). The effectiveness of the proposed adversarial loss is proven by demonstrating that the results surpass current state-of-the-art results.

In the experiment to validate the stochastic subject-wise mechanism method, we fixed the number of subjects used for meta-adapting and experimented by changing the number of subjects used for meta-training. Given that the meta-adapting set undergoes updates with T subjects in each main epoch, we opted to keep the number of subjects p in the meta-adapting set at a low value of 2 to minimize the  redundancy of subjects. As for the meta-training set, we conducted experiments with 5, 8, and 10 subjects k.
First, in the result without adversarial loss, the performance improved on the MPIIGaze dataset when k = 5, 8, and 10 compared to 4.8\degree (k[5]=6.8\%, k[8]=9.7\%, and k[10]=8\% improvement). In addition, all performances improved in the EyeDiap dataset compared to 6.0\degree (k[5]=21\%, k[8]=18.1\%, and k[10]=14,7\% improvement), all of which outperformed the current state-of-the-art (5.3\degree). Second, in the results applied adversarial loss, mean angle error results without the meta-process were 4.03\degree in the MPIIGaze and 5.02\degree in EyeDiap dataset. When the meta-process was applied, the performance improved for the MPIIGaze dataset when k = 8 and 10 (k[8]=3.5\%, k[10]=1.5\% improvement), and all performances improved for the EyeDiap dataset (k[5]=12\%, k[8]=10\%, k[10]=5.8\% improvement).

\begin{figure}[h]

\begin{description}
\footnotesize
\centering
\item \textbf{MPIIGaze } \\
\end{description}
\centering
\includegraphics[width=12cm]{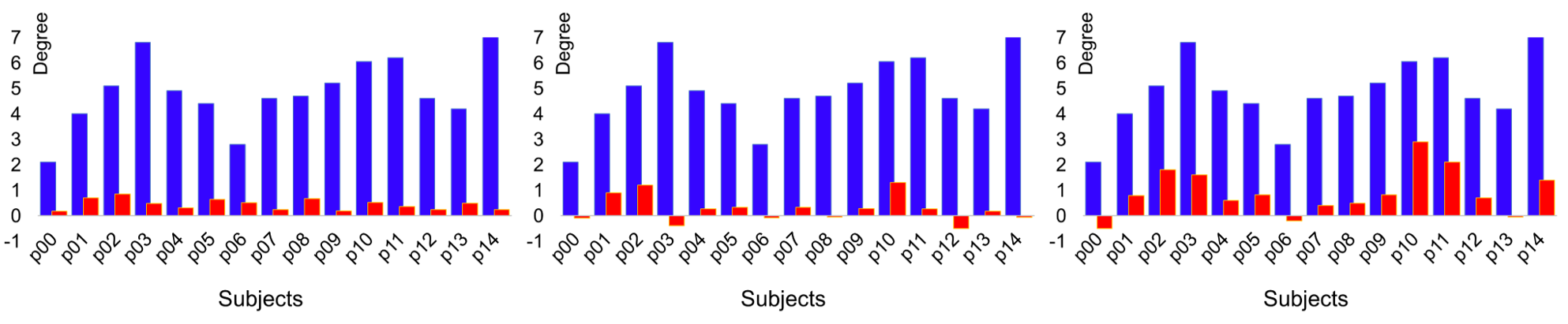}
\begin{description}
\footnotesize
\centering
\item \textbf{EyeDiap } \\
\end{description}
\centering
\includegraphics[width=12cm]{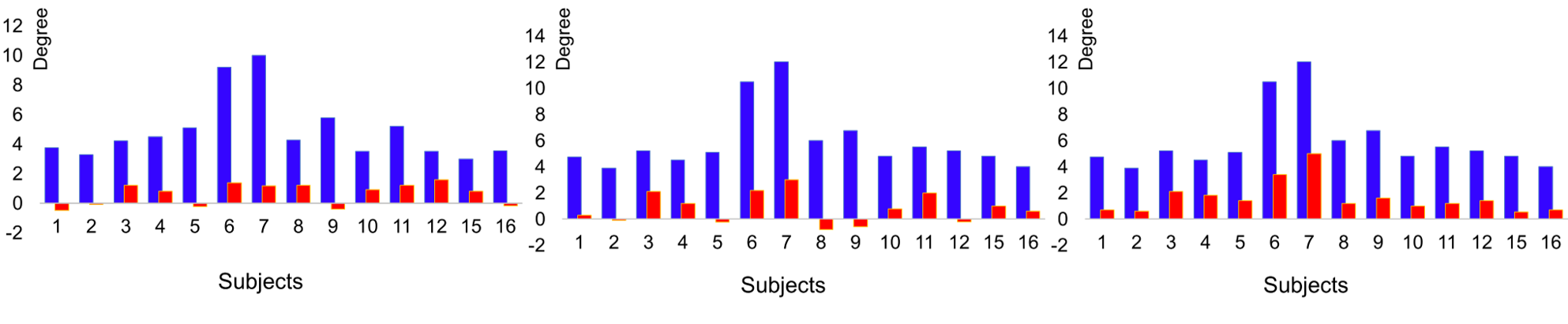}

\begin{description}
\footnotesize
\item \textbf{\quad(a) Adversarial loss \qquad\qquad(b) Subject-wise \qquad\qquad\quad (c) Both} 
\vspace{-0.2cm}
\item \textbf{ \qquad\qquad\qquad\qquad\qquad\qquad\qquad\quad optimization} 
\vspace{-0.4cm}
\end{description}
\caption{Improvement in gaze accuracy based on subject-wise optimization and adversarial loss respectively (the blue bars denote the baseline performance, while the red bars signify the observed improvement). (a) Improvement result based on adversarial loss. (b) Improvement for each subject based on subject-wise optimization. (c) Improvement for the combined application of both techniques.}
% \label{fig:example}
\end{figure}

To confirm the effects of adversarial loss and subject-wise optimization in detail, we analyzed the performance improvement for each subject when each method was applied individually. The performance improvement results for each subject are illustrated in Fig. 7. Fig. 7(a) shows the improvement when only adversarial loss is used, indicating that the performance was improved for all numbers of subjects on MPIIGaze,  whereas performance improvements were observed in subjects except 1, 2, 5, 9, and 16 for EyeDiap. Fig. 7(b) shows the improvements when the subject-wise optimization is used, indicating that a performance improvement was observed in subjects except p00, p03, p06, p08, p12 and p14 on MPIIGaze, whereas improvements were observed for all subjects except 2, 5, 8, 9, and 12 on EyeDiap. Generally, the overall difference in performance improvement was not high when only adversarial loss was applied (Fig. 7(a)), whereas the results using only subject-wise optimization (Fig. 7(b)) indicate that there was a higher difference in performance improvement for each subject. Regarding EyeDiap (Fig. 7(b)), the maximum performance improvement of subject 3 improved by 3\degree, whereas the maximum performance improvement shown in Fig. 7(a) was 1.38\degree. Similarly, regarding the MPIIGaze results, it was confirmed that the maximum performance improvement when using subject-wise optimization (Fig. 7(b)) was higher than that obtained by  adversarial loss (Fig. 7(a)). 
Finally, drawing in Fig. 7(c), the results from the combination of the two techniques show improved performance for all subjects in both MPIIGaze and EyeDiap datasets, except for subject p00 in MPIIGaze. The experimental results demonstrate that the combination of the proposed adversarial loss and subject-wise optimization provides positive effects on appearance-based gaze estimation.

\subsection{Gaze Distribution of The Generated Images}

\begin{figure}[h]
\centering
% figure6
\includegraphics[width=12cm]{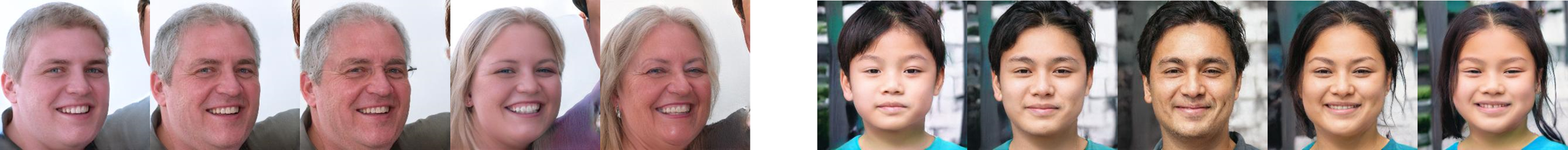}
\begin{description}
\centering
\footnotesize
\item \textbf{ Class 1\quad\qquad\qquad\qquad\quad\qquad\qquad\qquad\quad Class 2} \\
\end{description}

\includegraphics[width=12cm]{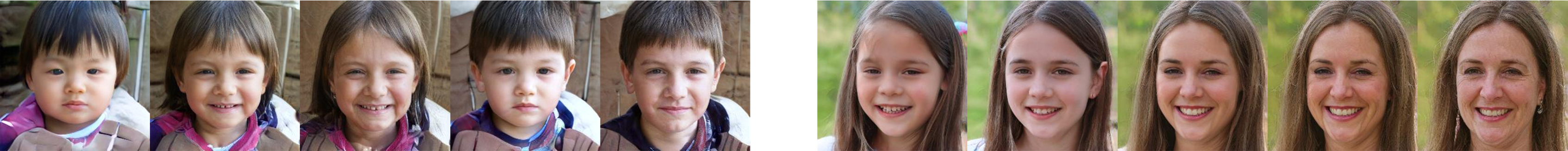}
\begin{description}
\centering
\footnotesize
\item \textbf{ Class 3\quad\qquad\qquad\qquad\qquad\quad\qquad\qquad\quad Class 4} \\
\end{description}

\includegraphics[width=12cm]{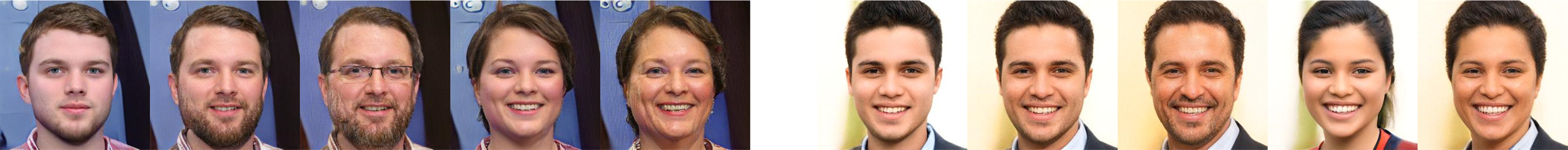}
\begin{description}
\centering
\footnotesize
\item \textbf{ Class 5\quad\qquad\qquad\qquad\qquad\quad\qquad\qquad\quad Class 6} \\
\end{description}

\includegraphics[width=6cm]{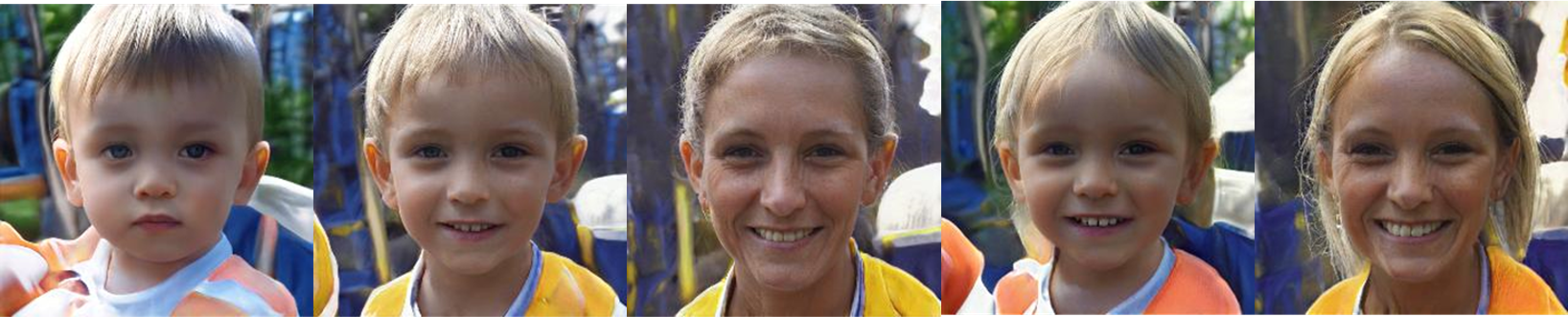}
\begin{description}
\centering
\footnotesize
\item \textbf{Class 7} \\
\end{description}

\vspace{-0.1cm}
\caption{Example of generated images of seven classes used for generalization effect verification. The images for each class is a set of different appearances with the similar gaze.}
\label{fig:short}
\end{figure}

\begin{figure*}
\centering
\includegraphics[height=8cm, width=12cm]{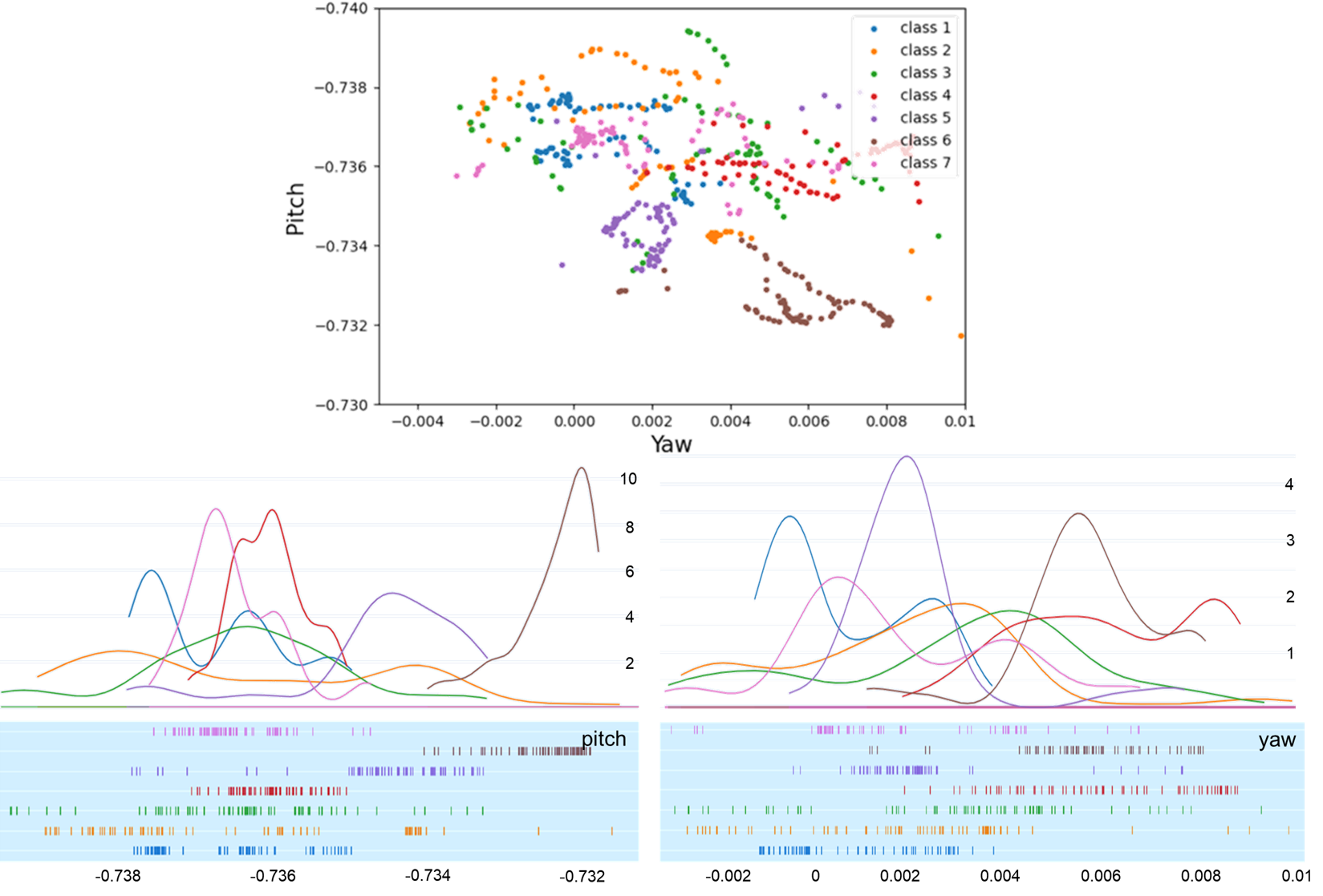}
\begin{description}
\centering
\footnotesize
\item \textbf{(a) Direction values predicted by Spatial-weight CNN with normal training} 
\end{description}

\includegraphics[height=8cm, width=12cm]{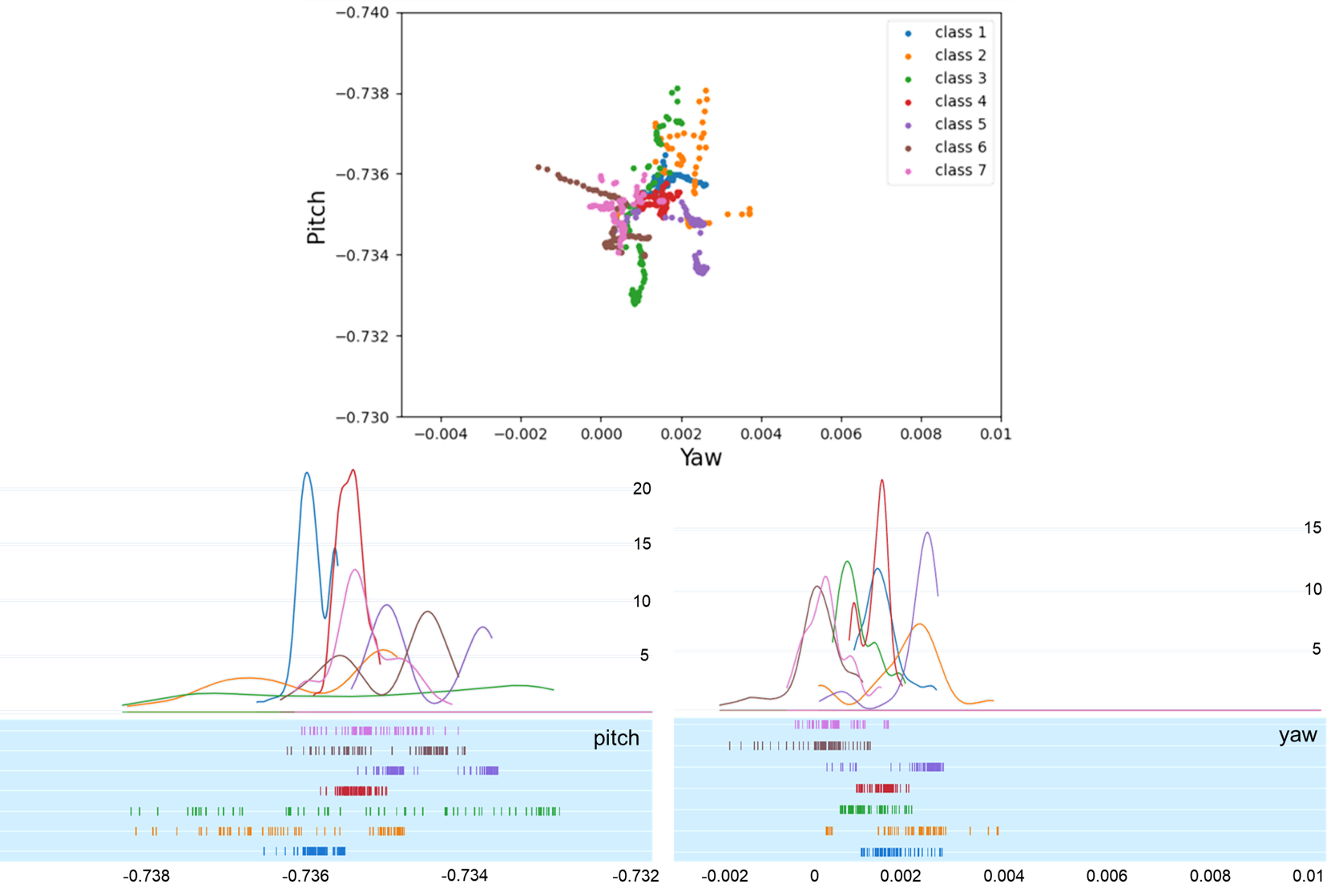}
\begin{description}
\centering
\footnotesize
\item \textbf{(b) Direction values predicted by Spatial-weight CNN with our proposed method}
\end{description}
\vspace{-0.1cm}
\caption{Evaluation of the generalization performance. Each class denotes images of different styles with the same gaze, generated from the generative model. The upper side is a scatter plot of yaw and pitch values, and the low side is the statistical representation of the yaw and pitch values.}
\label{fig:short}
\end{figure*}

For further evaluation, we conducted experiments using images generated from the generative model \cite{shen2020interpreting}. As mentioned in Section 4.3, we constructed a new dataset containing the seven gaze types and verify the generalization effect. Fig. 8 shows examples of some images created for the experiment. We settle on a total of seven reference images using InterFaceGAN \cite{shen2020interpreting} and changed latent distance about age, gender to obtain 70 different appearance images each with a similar gaze. Since the gaze value of the obtained image is unknown, we infer the gaze of the images from the model and check the distribution. For  Spatial-weight CNN as a baseline, Fig. 9 indicates the distribution of yaw and pitch values inferring from the constructed datasets, where Fig. 9(a) shows results for the training model without adversarial loss on meta-processing, and Fig. 9(b) shows the results after applying the proposed method. The upper side shows the results of displaying the predicted yaw and pitch values from all images of each class on the coordinate axis, and the low side shows the statistical representation, as named histogram, of the yaw and pitch. This result shows the formation of better clusters after applying our proposed method, indicating that our framework improves the generalization performance involving appearance factors in new environments. Table 3 quantifies the yaw and pitch variance for each class in more detail, indicating that our framework produces a lower variance in yaw and pitch. These evaluations show that our method is less biased toward a subject’s appearance factors and is robust to images obtained from various environments.

\begin{table}[h]
\begin{center}
\caption{The variance of yaw and pitch on images generated by the generative model. We compared the results by predicting the yaw and pitch of the image generated from the model trained using our method and the model not used. }
{\footnotesize
\begin{tabular}{c c c}
\hline
\multicolumn{1}{c}{} & \multicolumn{1}{c}{Normal training} & \multicolumn{1}{c}{Ours}\\
\hline\hline
\multirow{2}{*}{Class 1} & yaw : 2.1$\times${$10^{-6}$}&yaw : 1.6$\times${$10^{-7}$}\\

& pitch : 7.5$\times${$10^{-7}$} & pitch : 4.4$\times${$10^{-8}$}\\

\hline

\multirow{2}{*}{Class 2} & yaw : 7.0$\times${$10^{-6}$}&yaw : 6.8$\times${$10^{-7}$}\\

& pitch : 4.0$\times${$10^{-6}$} & pitch : 9.6$\times${$10^{-7}$}\\

\hline

\multirow{2}{*}{Class 3} & yaw : 8.5$\times${$10^{-6}$}&yaw : 1.4$\times${$10^{-7}$}\\

& pitch : 1.7$\times${$10^{-6}$} & pitch : 2.8$\times${$10^{-7}$}\\
\hline

\multirow{2}{*}{Class 4}& yaw : 3.6$\times${$10^{-6}$}&yaw : 6.8$\times${$10^{-8}$}\\
& pitch : 2.1$\times${$10^{-7}$} & pitch : 3.8$\times${$10^{-8}$}\\
\hline

\multirow{2}{*}{Class 5}& yaw : 2.6$\times${$10^{-6}$}&yaw : 3.2$\times${$10^{-7}$}\\
& pitch : 1.3$\times${$10^{-6}$} & pitch : 3.6$\times${$10^{-7}$}\\
\hline

\multirow{2}{*}{Class 6}& yaw : 2.4$\times${$10^{-6}$}&yaw : 3.2$\times${$10^{-7}$}\\
& pitch : 2.8$\times${$10^{-7}$} & pitch : 4.0$\times${$10^{-7}$}\\
\hline

\multirow{2}{*}{Class 7}& yaw : 5.4$\times${$10^{-6}$}&yaw : 1.9$\times${$10^{-7}$}\\
& pitch : 3.5$\times${$10^{-6}$} & pitch : 1.7$\times${$10^{-7}$}\\
\hline
\end{tabular}}
\end{center}
\end{table}

\subsection{Cross-dataset Evaluation}

\begin{table}
\begin{center}
\caption{
Results of cross-dataset gaze estimation performance compared to the baseline model and the existing gaze estimation frameworks. The bold texts denote the best performance except for SAZE+PnP-GA.}
{\footnotesize
\begin{tabular}{c c c c c c || c}
\hline
\multirow{2}{*}{} & \multirow{2}{*}{SWCNN} & \multirow{1}{*}{SWCNN} &\multirow{1}{*}{PARKS} &\multirow{1}{*}{ETH}& \multirow{1}{*}{SAZE}&\multirow{1}{*}{SAZE}\\
&  & \multirow{1}{*}{+ PnP-GA}  & \multirow{1}{*}{-Gaze} &\multirow{1}{*}{-XGaze}& \multirow{1}{*}{(ours)}& \multirow{1}{*}{+ PnP-GA}
% \multirow{3}{*}{} & SWCNN & SWCNN & SAZE\\
% & \cite{zhang2017s} & + PnP-GA  & (ours)
\\
% &&\cite{liu2021generalizing}&\\
\hline\hline
\multirow{2}{*}{$D_E \rightarrow D_M$} & \multirow{2}{*}{10.94$\degree$} & \multirow{2}{*}{8.14$\degree$} &\multirow{2}{*}{\textbf{6.6\degree}} & \multirow{2}{*}{7.5$\degree$} &\multirow{2}{*}{8.01$\degree$} &\multirow{2}{*}{5.89$\degree$}\\
&&&&&\\
\hline
\multirow{2}{*}{$D_E \rightarrow D_D$} & \multirow{2}{*}{24.95$\degree$} & \multirow{2}{*}{15.794$\degree$} & \multirow{2}{*}{20.2\degree}&\multirow{2}{*}{\textbf{11.0\degree}}&\multirow{2}{*}{12.9$\degree$}&\multirow{2}{*}{10.3$\degree$}\\
&&&&&\\
\hline
\multirow{2}{*}{$D_G \rightarrow D_M$} & \multirow{2}{*}{10.02$\degree$} &\multirow{2}{*}{\textbf{8.74$\degree$}} & \multirow{2}{*}{-}&\multirow{2}{*}{10.3\degree}&\multirow{2}{*}{9.89$\degree$}&\multirow{2}{*}{6.23$\degree$}\\
&&&&&\\
\hline
\multirow{2}{*}{$D_G \rightarrow D_D$} & \multirow{2}{*}{13.47$\degree$} &\multirow{2}{*}{11.38$\degree$} & \multirow{2}{*}{-}&\multirow{2}{*}{11.3\degree}&\multirow{2}{*}{\textbf{9.05$\degree$}}&\multirow{2}{*}{7.86$\degree$}\\
&&&&&\\
\hline

\multirow{2}{*}{$D_{GC} \rightarrow D_M$} & \multirow{2}{*}{7.23\degree} &\multirow{2}{*}{-} & \multirow{2}{*}{5.3\degree}&\multirow{2}{*}{4.5\degree}&\multirow{2}{*}{\textbf{4.32}}&\multirow{2}{*}{4.13$\degree$}\\
&&&&&\\
\hline

\multirow{2}{*}{$D_{GC} \rightarrow D_D$} & \multirow{2}{*}{21.60} &\multirow{2}{*}{-} & \multirow{2}{*}{19.0\degree}&\multirow{2}{*}{13.7\degree}&\multirow{2}{*}{\textbf{10.4}}&\multirow{2}{*}{8.12$\degree$}\\
&&&&&\\
\hline

\end{tabular}}
\end{center}

\end{table}

To verify the effectiveness of our proposed SAZE, we conducted a cross-dataset evaluation with several gaze estimation methods. For the experiment, the first three datasets ETH-XGaze \cite{zhang2020eth} $(D_E)$, Gaze360 \cite{kellnhofer2019gaze360} $(D_G)$ and The GazeCapture dataset \cite{krafka2016eye} $(D_{GC})$ were processed using the proposed SAZE framework and then tested using \cite{zhang2017s} $(D_M)$ and EyeDiap \cite{funes2014eyediap} $(D_D)$ to evaluate the cross-dataset gaze estimation performance. ETH-XGaze provides 80 subjects (756,540 images) and includes data labels for training. The Gaze360 dataset consists of full-body and close-up head images. To train using Gaze360, we removed the full-body images and cropped all faces from the closer-up head images, resulting in 100,933 images. The GazeCapture dataset comprises data from a total of 1,474 participants, encompassing a substantial collection of 2.4 million images. In light of the absence of 3D gaze annotations in the GazeCapture dataset, we leveraged the normalized gaze annotation by \cite{park2019few}. Our dataset was partitioned into three sets, comprising 1,176 participants for training, and 140 for the test split. In previous studies, appearance-based gaze estimation has primarily focused on utilizing datasets composed of still images rather than video sequences. Consequently, these specific datasets, consisting of video sequences, have not been extensively employed as the primary dataset for appearance-based gaze estimation in previous research. As a result, our paper employs these datasets solely for the purpose of validating the effectiveness of adaptation. For validation, training on the ETH-XGaze dataset was performed with 80 subjects with train subject k = 30 and adapt subject p = 15, and training on the Gaze Capture dataset was performed with k = 200 and p = 50 out of 1,176 subjects. Next, unlike other datasets, the Gaze360 dataset is divided into 80 folders, each consisting of 1-3 different subjects. Therefore, we set k = 30 and p = 15 for training based on one folder.

Table 4 leads to an insightful conclusion. Table 4 presents the results obtained from different implementation of the SWCNN \cite{zhang2017s} model, which serves as our baseline. The first column corresponds to the cross-dataset results of SWCNN, while the second column represents the outcomes after applying the plug-and-play gaze adaptation (PnP-GA) framework to SWCNN, as described in \cite{liu2021generalizing}. In the second column, the PnP-GA framework, is a domain adaptation technique that that leverages a small number of target domain samples for adaptation. The third column displays the evaluation results on the cross-dataset, as presented in the paper introducing the PARKS gaze dataset \cite{murthy2023towards}. Meanwhile, the fourth column showcases the outcomes from the research paper introducing the ETH-XGaze dataset \cite{zhang2020eth}.

In the results of the second column, the proposed method outperformed the gaze domain adaptation method \cite{liu2021generalizing} for all results except for the $D_G \rightarrow D_M$ case, even when we did not use an adaptation technology between the source and target domains. These results show that our proposed method is beneficial not only for generalization but also for domain adaptation. Upon close examination of the results in the third column PARKS-Gaze, it is evident that, with the exception of the results for $D_E \rightarrow D_M$, our method demonstrates commendable performance. Notably, in the case of the EyeDiap dataset, we observe a substantial improvement in performance across both scenarios. Furthermore, it is worth noting that PARKS-Gaze employed a model trained on both eye images and face image for cross-dataset evaluation. In contrast, our proposed method relies exclusively on face images for training. Despite this constraint, we have demonstrated superior performance over PARKS-Gaze in three cases. The outcomes presented in the fourth column ETX-XGaze unequivocally affirm our method's superiority on four cases. Unlike ETH-XGaze, which utilized ResNet50, we employed ResNet18, and it demonstrated strong performance across four cases. Finally, when comparing SWCNN with SAZE, we observed improvements in performance across all cases. Particularly, in case $D_E \rightarrow D_D$, SAZE demonstrated a significant performance enhancement of approximately 48\%.

Furthermore, to verify the compatibility of our method with the domain adaptation framework, we tried to perform a cross-dataset evaluation by combining PnP-GA with SAZE. The results show that the performance is better than using only a single SAZE in all parts, which can be seen in the last column. This demonstrates the compatibility of our method with the adaptation framework.

%% file: Writing/6_conclusion.tex
\section{Conclusion}
In this paper, we present a novel Stochastic subject-wise Adversarial GaZE learning framework (SAZE) applicable to appearance-based gaze estimation. The proposed adversarial loss makes it possible to extract appearance-independent gaze representations, whereas subject-wise optimization alleviates the bias caused by the limited number of subjects generally associated with gaze datasets. Our method demonstrated superior gaze accuracy on two public datasets. The main goal of this study is the generalization of appearance factors. Because gaze estimation is used as an important cue for many applications, we propose SAZE, which addresses the inevitable limitations of gaze estimation associated with appearance factors. 

However, we must acknowledge certain limitation in our study. A closer examination of the results presented in Figure 7 in Section 5.4 reveals that the adversarial loss demonstrates effectiveness on the subjects generally. Conversely, subject-wise optimization yields unfavorable outcomes for some specific subjects. In order to mitigate any potential bias towards appearance in a data environment with a limited subject pool, we meticulously organized both the meta-training set and the meta-adapting set in a configuration where subjects do not overlap, and subsequently conducted the meta-process. Nevertheless, we conducted empirical experiments to determine the optimal number of subjects in each set to achieve superior performance. Additionally, during the construction of the meta-adapting set, we did not rely on any statistical analysis, but rather randomly selected subjects for T steps. We firmly believe that for a more reliable application of the meta-process, a more thorough and statistically grounded analysis is imperative to ascertain the optimal conditions for any given dataset. This constitutes a pivotal area for future improvement, one that will undoubtedly bolster the robustness of our method. We leave the evaluation of this requirement as a future work. Notwithstanding these constraints, we demonstrated the generalization effect obtained by SAZE in the expanded experimental results, demonstrating possibilities for application scalability with respect to images taken in different environments.

Moreover, traditional studies commonly incorporate additional information, such as utilizing both eye images and a face image, or implementing complex model architectures, to enhance performance. In our comparative analysis with AGE-Net, it was observed that the AGE-Net, as implemented based on their published documentation, possesses approximately 1.6 million parameters. In contrast, the network utilized in our experiments comprises 2.8 million parameters. However, AGE-Net, which incorporates both eye and facial images, exhibits Floating Point Operations Per Second (FLOPs) of 5.35 billion. Conversely, our methodology, which relies solely on single facial images, yielded substantially lower FLOPs, registering at 1.42 billion. This significant reduction in computational complexity underscores the efficiency of our approach while maintaining robust performance metrics. Also, domain adaptation studies typically rely on target domain templates to ensure broader generalization. In contrast, our approach employs a minimalist approach, utilizing solely ResNet18 and face images, yet yields improvements in both performance and generalization, all without relying on target domain. Based on these results, we are confident that our proposed framework will help in future gaze estimation studies.\\

\noindent $\textbf{Acknowledgement}$ This work was supported by Institute of Information \& communications Technology Planning \& Evaluation (IITP) grant funded by the Korea government (MSIT) (No. 2019-0-00079, Artificial Intelligence Graduate School Program (Korea University) and No. 2022-0-00984, Development of Artificial Intelligence Technology for Personalized Plug-and-Play Explanation and Verification of Explanation).